\documentclass[10pt,twocolumn,twoside]{IEEEtran}
\usepackage{graphicx,subfigure}
\usepackage{epstopdf}
\usepackage{amsmath}
\usepackage{amssymb}
\usepackage[linesnumbered,ruled,vlined]{algorithm2e}
\usepackage{cite}
\usepackage{bm}
\usepackage{color}
\numberwithin{equation}{section}

\DeclareMathOperator*{\argmin}{arg\,min}

\def\ben{\begin{equation*}}
\def\een{\end{equation*}}
\def\be{\begin{equation}}
\def\ee{\end{equation}}
\def\beaa{\begin{eqnarray*}}
\def\eeaa{\end{eqnarray*}}
\def\bea{\begin{eqnarray}}
\def\eea{\end{eqnarray}}

\ifCLASSINFOpdf
\else
\fi
\begin{document}
%
\title{Structured Dictionary Learning for Classification}
%
%
%

\author{Yuanming~Suo,~\IEEEmembership{Student Member,~IEEE,}
        Minh~Dao,~\IEEEmembership{Student Member,~IEEE,}
        Umamahesh~Srinivas,~\IEEEmembership{Student Member,~IEEE,}
        Vishal~Monga,~\IEEEmembership{Senior Member,~IEEE,}
        and~Trac~D.~Tran,~\IEEEmembership{Fellow,~IEEE}
\thanks{Y. Suo, M. Dao, T. D. Tran are with The Johns Hopkins University, Baltimore,
MD, 21218 USA. (email: ysuo1@jhu.edu.)}
\thanks{U. Srinivas and V. Monga are with The Pennsylvania State University, University Park, USA.}
\thanks{This work has been partially supported by NSF under Grant CCF-1117545, ARO under Grant 60219-MA, and ONR under Grant N00014-12-1-0765.}}

%
%

\markboth{submitted to IEEE TRANSACTIONS ON SIGNAL PROCESSING}%
{Shell \MakeLowercase{\textit{et al.}}: Bare Demo of IEEEtran.cls for Journals}
%



\maketitle

\begin{abstract}
Sparsity driven signal processing has gained tremendous popularity in the last decade. At its core, the assumption is that the signal of interest is sparse with respect to either a fixed transformation or a signal dependent dictionary. To better capture the data characteristics, various dictionary learning methods have been proposed for both reconstruction and classification tasks. For classification particularly, most approaches proposed so far have focused on designing explicit constraints on the sparse code to improve classification accuracy while simply adopting $l_0$-norm or $l_1$-norm for sparsity regularization. Motivated by the success of structured sparsity in the area of Compressed Sensing, we propose a structured dictionary learning framework (StructDL) that incorporates the structure information on both group and task levels in the learning process. Its benefits are two-fold: \textit{(i)} the label consistency between dictionary atoms and training data are implicitly enforced; and \textit{(ii)} the classification performance is more robust in the cases of a small dictionary size or limited training data than other techniques. Using the subspace model, we derive the conditions for StructDL to guarantee the performance and show theoretically that StructDL is superior to $l_0$-norm or $l_1$-norm regularized dictionary learning for classification. Extensive experiments have been performed on both synthetic simulations and real world applications, such as face recognition and object classification, to demonstrate the validity of the proposed DL framework.
\end{abstract}


%
\IEEEpeerreviewmaketitle

\section{Introduction}
%
%
%
%
\IEEEPARstart{I}{n} many areas across science and engineering, researchers are dealing with signals that are often inherently sparse with respect to a certain dictionary (also called basis or transform). The seminal paper  by neuroscientists Olshausen and Field \cite{olshausen1997sparse} points out that the receptive fields in human being's visual cortex utilize sparse coding to extract meaningful information from images. In the signal processing domain, the emerging field of Compressed Sensing (CS) \cite{candes2006robust} relies on the key assumption that the signal is sparse under some orthogonal transformations, such as the Fourier transform. 

Traditionally, dictionaries are designed for desired properties in spatial or frequency domain or both. Recently, a different methodology to learn the dictionary from data is explored, which could better capture data characteristics. There are two different directions for designing such a signal dependent dictionary:
\newline
\textit{(i)} Using data directly as the dictionary: Wright \textit{et al.} \cite{wright2009robust} proposed a sparse representation-based classifier (SRC) that concatenates the training data from different classes into a single dictionary and uses class-specific residue for face recognition. Besides supervised tasks, a data dictionary is also  utilized to cluster the high dimensional data by finding intrinsic low dimensional structures with respect to itself \cite{elhamifar2012sparse}.
\newline
\textit{(ii)} Training a dictionary using data: Aharon \textit{et al.} \cite{aharon2006img} proposed an algorithm called K-SVD that guarantees all training data to be sparsely represented by the learned dictionary and demonstrated its advantages in image processing tasks. Yu \textit{et al.} \cite{yu2009nonlinear} justified that encoding data with dictionary atoms in its neighborhood can guarantee a nonlinear function of the data to be well approximated by a linear function. 
 
In contrast to the former approach, the learned dictionary in the latter approach removes the redundant information in the learning process, therefore the size of the dictionary does not grow with the size of the data. In this paper, we will focus on the latter approach. Moreover, we assume that the data has been properly aligned, although data alignment \cite{shekhar2013generalized, qiu2013learning} is another active research area with growing interests.

\subsection{Dictionary Learning for Reconstruction}
Dictionary learning (DL) is first attempted for the purpose of reconstruction. The learning process can be described by following optimization problem:
\begin{equation*} \label{eq:dictmodel1}
\min_{\mathbf{D}, \mathbf{A}}\sum_{i=1}^{N}{(\frac{1}{2}||\mathbf{x}_i - \mathbf{D} \mathbf{a}_i||_2^2 + \lambda_1 ||\mathbf{a}_i||_q)}.
\end{equation*}
Given training data $\mathbf{x}_i \in \mathbb{R}^M$ ($i=1,...,N$), the dictionary $\mathbf{D} \in \mathbb{R}^{M \times K}$ and corresponding sparse coefficients $\mathbf{A} \in \mathbb{R}^{K \times N}$ are both learned. Each column of $\mathbf{D}$ and $\mathbf{A}$ are denoted as $\mathbf{d}_j$ ($j=1,...,K$) and $\mathbf{a}_i$ ($i=1,...,N$), respectively. The dictionary size $K$ is typically larger than signal dimension $M$. The parameter $\lambda_1$ balances the trade-off between data fidelity and the sparsity regularization via the $l_q$-norm. 

This non-convex optimization problem is usually solved by iterating between sparse coding and dictionary updating. In the sparse coding stage, the sparse coefficient $\mathbf{a}_i$ is found with respect to a fixed dictionary $\mathbf{D}$. This can be carried out by greedy pursuit enforcing constraints on $l_0$-norm \cite{aharon2006img}, convex optimization targeting $l_1$-norm \cite{mairal2009online,ramirez2012mdl}, minimizing $l_2$-norm with locality constraint \cite{yu2009nonlinear}, optimizing structured sparsity \cite{jenatton2010proximal, zelnik2012dictionary} or Bayesian methods \cite{zhou2012nonparametric}. 
In the dictionary updating stage, each dictionary atom $\mathbf{d}_j$ is updated using only data with non-zero sparse coefficients on index $j$. This sub-problem can be solved by either block coordinate descent \cite{mairal2009online} or singular value decomposition \cite{aharon2006img}. Desirable features, such as multi-resolution \cite{mairal2007learning} and transformation invariant \cite{kavukcuoglu2009learning}, could also be integrated to further improve performances in specific applications. Note that all the dictionary atoms should have unit $l_2$-norm to avoid the scenario that dictionary atoms have arbitrary large norm but sparse codes have small values.  

\subsection{Dictionary Learning for Classification}
\begin{figure}
\centering
\includegraphics[width=3in]{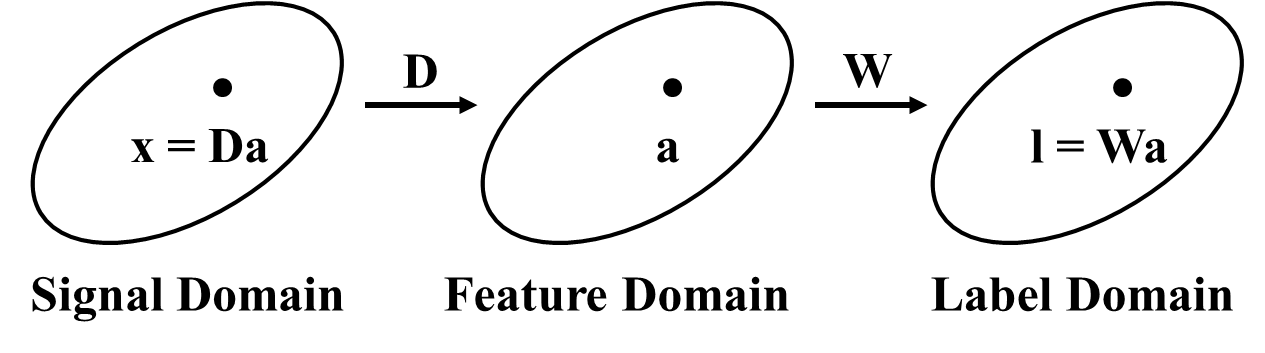} 
\caption{A schematic of using DL for classification.}
\label{fig:Concept}
\end{figure}

Notice that sparse coefficients could also be interpreted as features, therefore it is natural to explore the benefits of using DL for classification. A general framework for this purpose is illustrated in Fig~\ref{fig:Concept}. The low dimensional signal $\mathbf{x}$ is mapped to its high dimensional feature (sparse coefficient) $\mathbf{a}$ using a learned dictionary $\mathbf{D}$, which could make the hidden patterns more prominent and easier to capture. A classifier $\mathbf{W}$ is then utilized to predict the label vector $\mathbf{l}$. The key here is to design $\mathbf{D}$ and $\mathbf{A}$ with discriminative properties by adding extra constraints $f_{\mathbf{A}}(\cdot)$ and $f_{\mathbf{D}}(\cdot)$. Now the optimization problem becomes:
\begin{equation*} \label{eq:dictmodel2}
\min_{\mathbf{D}, \mathbf{A}}\sum_{i=1}^{N}{(\frac{1}{2}||\mathbf{x}_i - \mathbf{D} \mathbf{a}_i||_2^2 + \lambda_1 ||\mathbf{a}_i||_q) + \lambda_2 f_{\mathbf{A}}(\mathbf{A}) +\lambda_3 f_{\mathbf{D}}(\mathbf{D})}.
\end{equation*}
The function $f_{\mathbf{A}}(\cdot)$ could be a logistic function \cite{mairal2012task}, a linear classifier \cite{rodriguez2008sparse,zhang2010discriminative}, a label consistency term \cite{jiang2011learning, zhang2013online}, a low rank constraint \cite{zhang2013learning} or Fisher discrimination criterion \cite{yang2011fisher}. An example of $f_{\mathbf{D}}(\cdot)$ is to force the sub-dictionaries for different classes to be as incoherent as possible \cite{ramirez2010classification}. The label can be assigned using class-specific residue \cite{ramirez2010classification} or linear classification \cite{jiang2011learning}. Most aforementioned methods embed the label information into the DL problem explicitly, which could complicate the optimization procedure \cite{yang2011fisher}. 

\subsection{Our Contributions and Paper Structure}
Most methods mentioned in Section I.B simply add extra classification constraints on top of the DL formulation for reconstruction. In contrast to these approaches, we focus on improving the intrinsic discriminative properties of the dictionary by introducing a structured dictionary learning framework (StructDL) that incorporates structured sparsity on different levels. Our specific contributions are listed below\footnote{Preliminary version of this work will be presented at the IEEE International Conference on Image Processing, 2014 \cite{suo2014gddl}.}.
\begin{itemize}
\item In contrast to the approaches that add extra constraints\cite{zhang2010discriminative, jiang2011learning}, our formulation does not increase the size of the problem because the regularization is enforced implicitly. Different from approaches using group sparsity \cite{Yu-Tseh2013}, structured low rank \cite{zhang2013learning} and hierarchical tree sparsity constraints \cite{jenatton2010proximal} in DL, we propose to use hierarchical group sparsity, which can be naturally extended to its multi-task variation $-$ group structured dirty model for regularization. More importantly, the latter can uniquely incorporate sparsity, group structure and locality in a single formulation, which are all desired features for an ideal dictionary to be used in classification.
\item We show theoretically that our approach has the advantage of perfect block structure for classification at the cost of a stricter condition. We also point out that the condition is more likely to be satisfied when the dictionary size is smaller, thus making our method more favorable than $l_1$-norm based DL.
\item We employ both synthetic and real-world datasets to illustrate the superior performance of the proposed StructDL framework. Meanwhile, we also point out scenarios where limitations still exist.
\end{itemize}

The paper is organized as follows. In Section II, we illustrate the structured dictionary learning framework for classification (StructDL), including its single task and multi-task versions. In Section III, we derive conditions to guarantee its classification performance using a noiseless model. In Section IV, extensive experiments are performed with synthetic and real datasets to compare StructDL with other state-of-art methods. We end the paper with a conclusion and a discussion on future work in Section V.

\subsection{Notation}
In this section, we introduce notations that will be used throughout the article. We use bold lower-case letters such as $\mathbf{x}$ to represent vectors, bold upper-case letters such as $\mathbf{D}$ to represent matrices, and bold lower-case letter with subscript such as $\mathbf{d}_j$ to represent columns of a matrix. The dimensions of vectors and matrices are often clear from the context. For any vector $\mathbf{a}$, we use $||\mathbf{a}||_q$ to denote its $l_q$-norm $(0 \leq q \leq \infty)$. A group $g$ is a subset of indices in $\{1,...,K\}$. A group structure $\mathcal{G}$ denotes a pre-defined set of non-overlapping groups. We use $\rho(\cdot)$, $tr(\cdot)$, $rank(\cdot)$ and $dim(\cdot)$ to denote spectral norm, trace, rank of the matrix and dimension of the subspace, respectively.

\begin{figure*}[t]
\centering
\subfigure[$l_1$-norm based DL]{\includegraphics[width=1.58in]{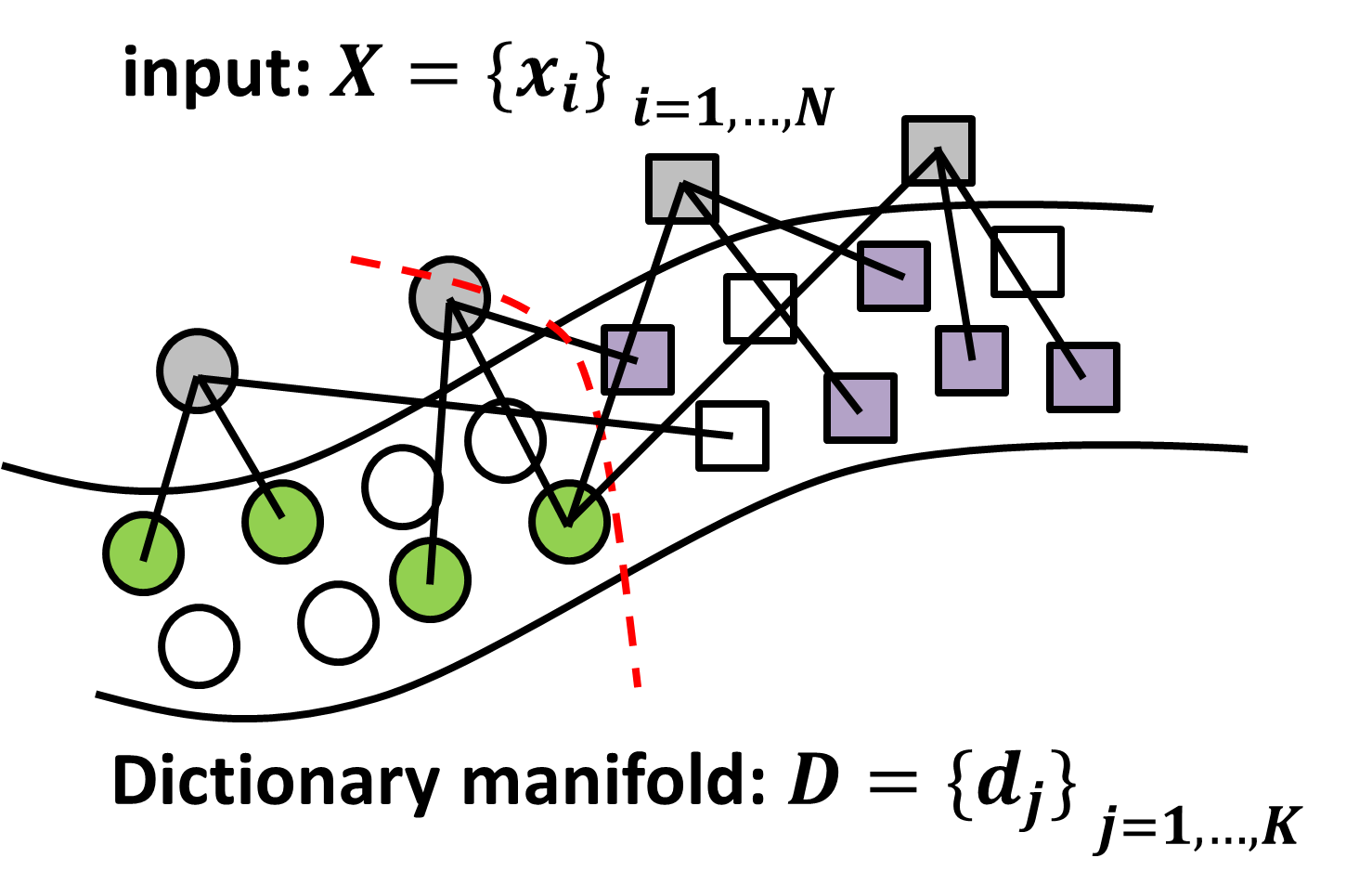}}
\subfigure[locality based DL]{\includegraphics[width=1.58in]{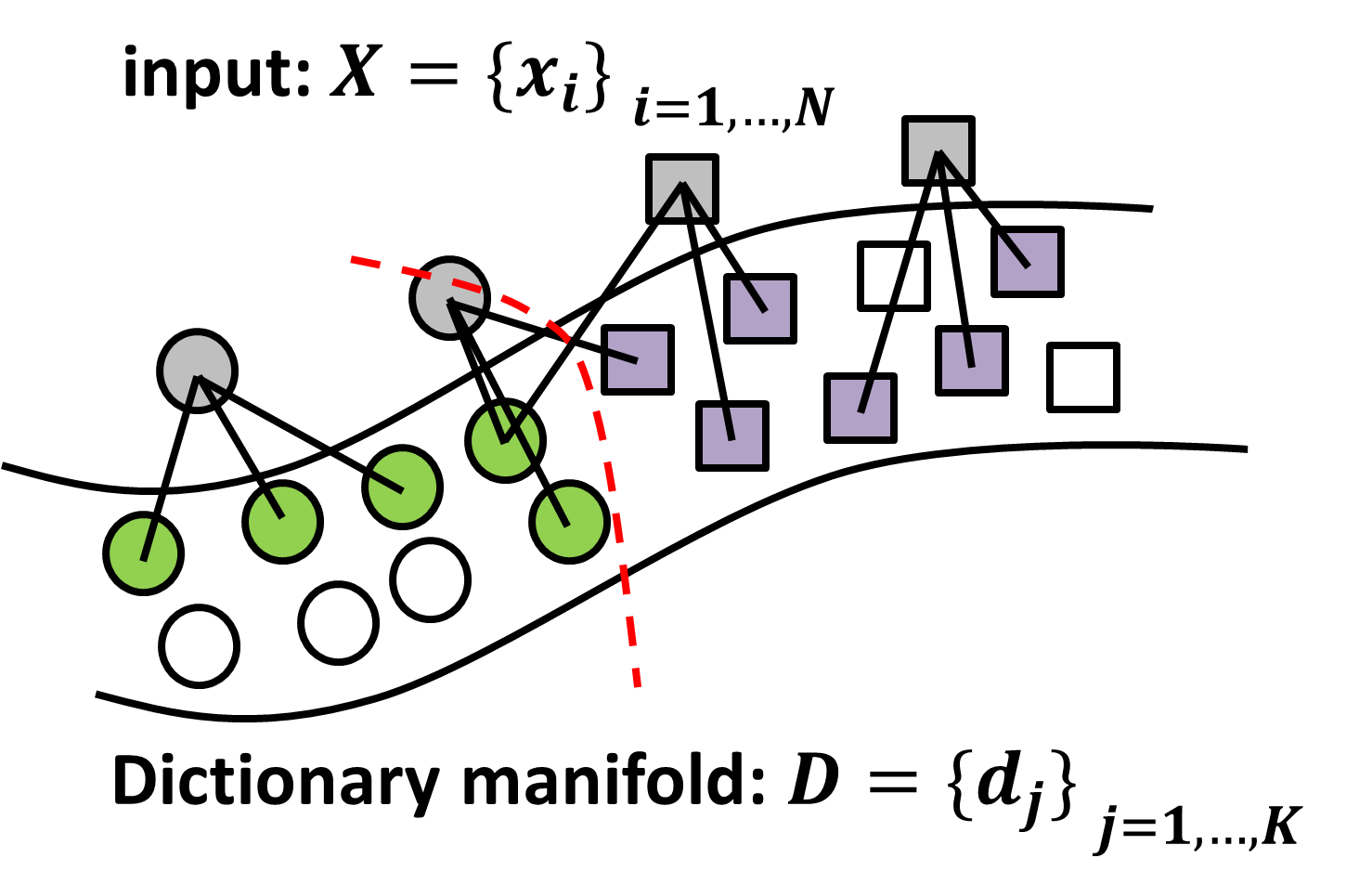}}
\subfigure[proposed HiDL]{\includegraphics[width=1.58in]{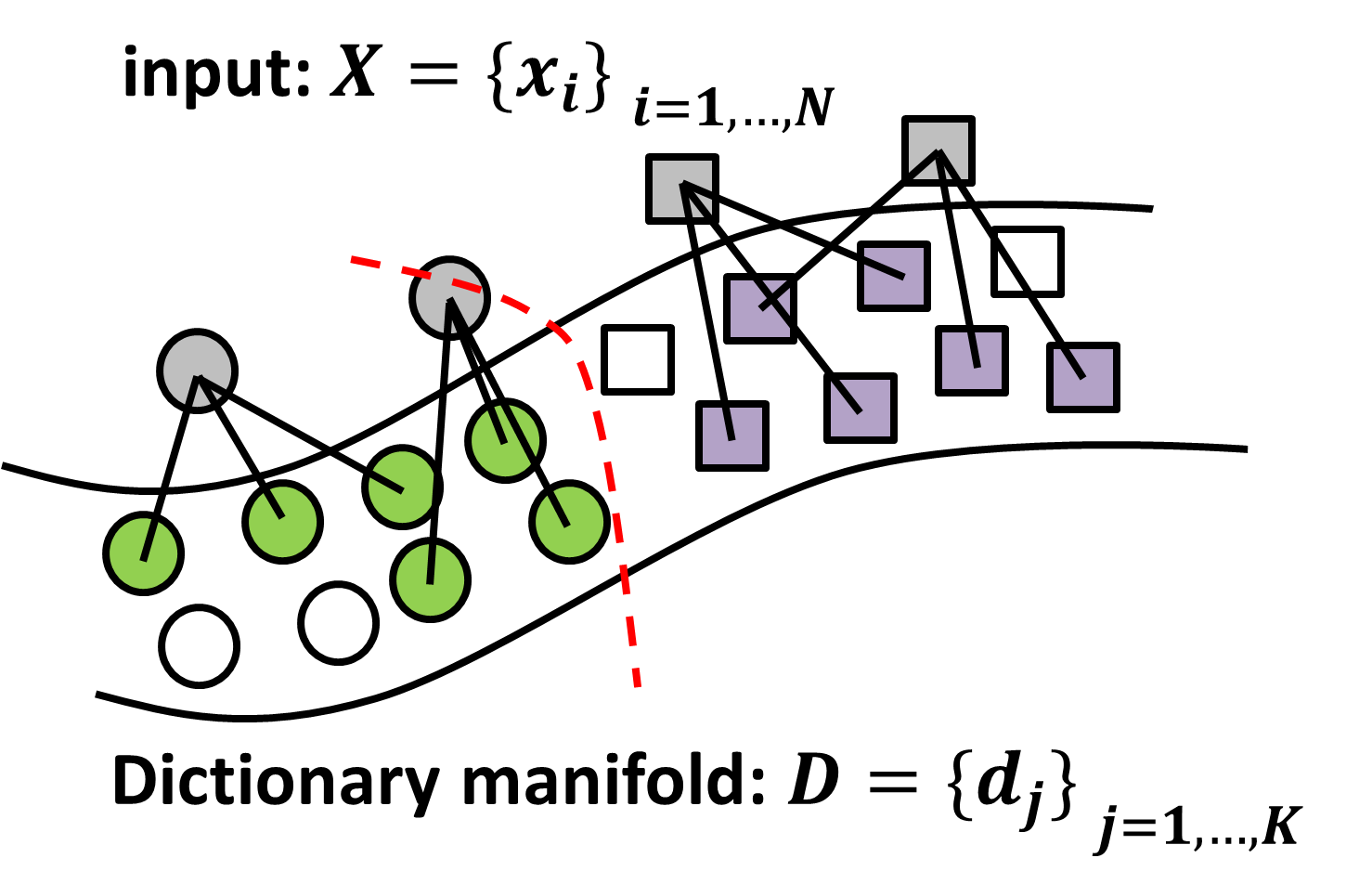}}
\subfigure[proposed GDDL]{\includegraphics[width=1.58in]{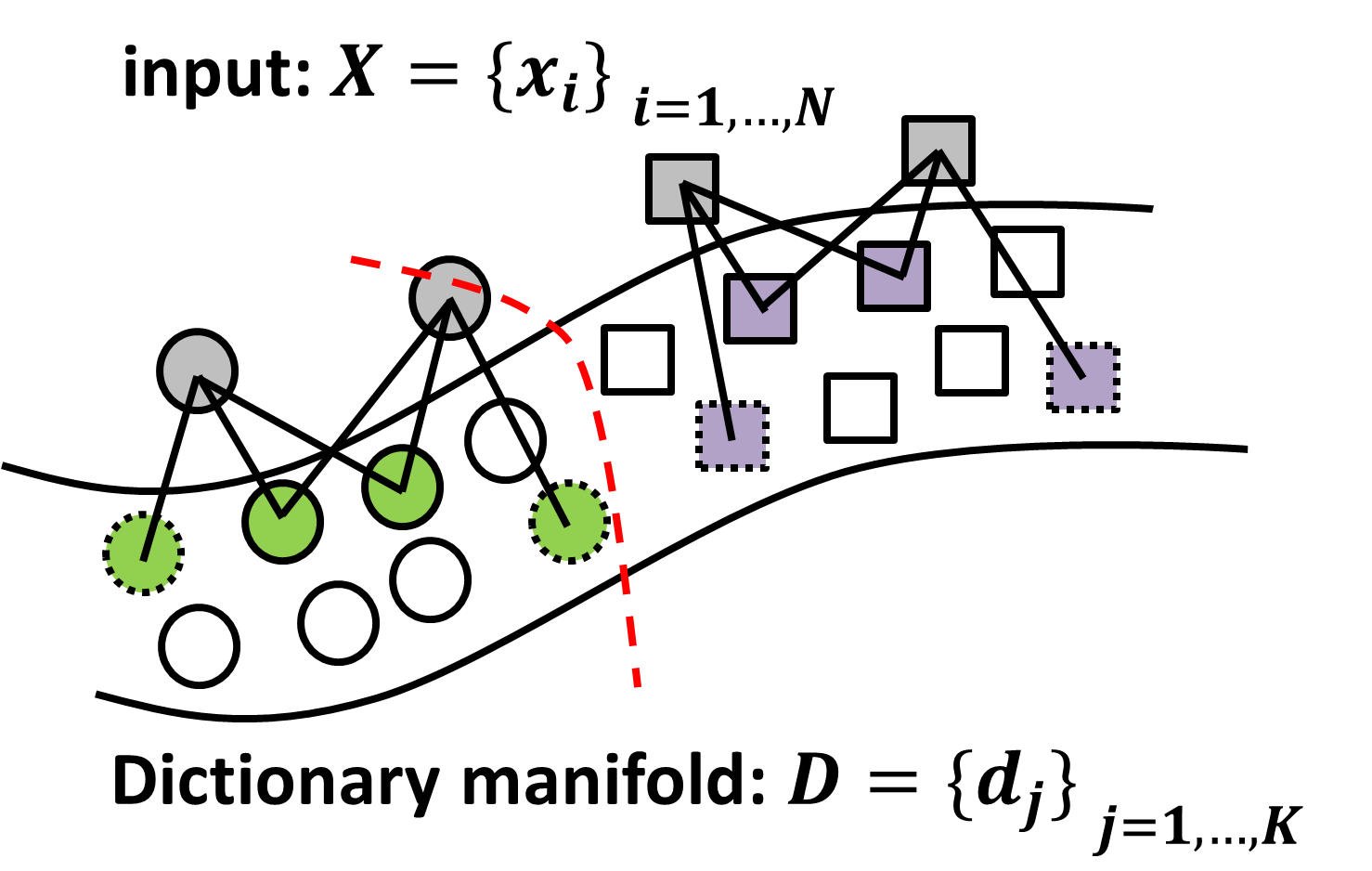}}
\caption{Comparison of two proposed StructDL approaches with other methods. Data matrix $\mathbf{X}$ are represented by grey circles and squares, corresponding to two different classes. The dictionary $\mathbf{D}$ lies on an oblique manifold \cite{absil2009optimization}. Green and purple indicates selected dictionary atoms from different classes. Red dotted curve represents the boundary that separates sub-dictionaries of different classes. In (a), $l_1$-norm based DL maps the data to a few dictionary atoms without limitation on their locations. In (b), the input is mapped to a few dictionary atoms in a certain neighborhood by locality constraint. However, data close to the class boundary could still be mapped to the dictionary atoms from wrong classes. In (c), HiDL forces the data to use a few atoms from same sub-dictionary (same class). In (d), GDDL separates the chosen atoms with the same label to two sub-groups: shared dictionary atoms (solid colored circle and square) and unique dictionary atoms (dashed colored circle and square).}
\label{fig:StructDL}
\end{figure*}

\section{Structured Dictionary Learning For Classification}
\subsection{Motivation from a Coding Perspective}
The coding stage in the DL process typically adopts $l_0$- or $l_1$-norm to encourage sparsity (the latter one is also referred as Lasso \cite{tibshirani1996regression}). Its formulation is
\begin{equation} 
\min_{\mathbf{A}}\sum_{i=1}^{N}{(\frac{1}{2}||\mathbf{x}_i - \mathbf{D} \mathbf{a}_i||_2^2 + \lambda_1 ||\mathbf{a}_{i}||_1}).
\end{equation}
The corresponding prior distribution for Lasso is a multivariate Laplacian distribution with the independence assumption, thus the chosen support could fall anywhere.

Since sparsity alone could not regulate the support location, locality-constrained linear coding (LLC) \cite{wang2010locality} is proposed to enforce locality instead of sparsity. The objective function of LLC is defined as:
\begin{equation}
\min_{\mathbf{A}}\sum_{i=1}^{N}{(\frac{1}{2}||\mathbf{x}_i - \mathbf{D} \mathbf{a}_i||_2^2 + \lambda_1||\mathbf{e}_i \odot \mathbf{a}_i||_2^2}),
\end{equation} 
where $\odot$ denotes the element-wise multiplication, and $\mathbf{e}_i \in \mathbb{R}^K$ is a weight vector indicating the similarity between signal and dictionary atoms. By controlling the size of the neighborhood, locality constraint could lead to sparsity as well. Conceptually, LLC endorses the local structure in the dictionary but loses the global perspective. For instance, the data lying on the class boundary could be coded with dictionary atoms from either side or both sides, creating ambiguity for classification tasks.

To promote both sparsity and group structure, Hierarchical Lasso (HiLasso) \cite{sprechmann2011c} is proposed as:
\begin{equation} \label{eq:hilasso}
\min_{\mathbf{A}}\sum_{i=1}^{N}{(\frac{1}{2}||\mathbf{x}_i - \mathbf{D} \mathbf{a}_i||_2^2 + \lambda_1 \sum_{g \in \mathcal{G}}^{}{||\mathbf{a}_{i,[g]}||_2}} + \lambda_2 ||\mathbf{a}_i||_1),
\end{equation}
where $\mathcal{G}$ is a predefined group structure, and $\mathbf{a}_{i,[g]}$ is the sub-vector extracted from $\mathbf{a}_{i}$ using the indices in group $g$. The group structure of HiLasso naturally yields locality because it reflects the clustering of dictionary atoms. It is also relevant for classification tasks, since this grouping of dictionary atoms naturally reflects their labels. To be more specific, the dictionary $\mathbf{D}$ is the concatenation of sub-dictionaries $\mathbf{D}_1,...,\mathbf{D}_C$ belonging to different classes, where $C$ is the total number of classes and $\mathbf{D}_c \; (c = 1,...,C)$ has size $K_c$. In contrast to LLC, HiLasso captures the global information embedded in the group structure.  

In the multi-task setup, different tasks could share same sets of dictionary atoms, which leads to a variant of HiLasso, called Collaborative HiLasso (C-HiLasso) \cite{sprechmann2011c}. C-HiLasso captures the correlation on the group level, but it does not reveal explicitly if any dictionary atoms are shared by all tasks (within-class similarity) or uniquely utilized by individual task (within-class variation). The within-class variation generally makes the data clusters less compact and harder to classify, therefore it will be beneficial to separate it from the within-class similarity component to better capture the core essence of the data for discriminative applications. A mixture of coefficients model is proposed to carry out this decomposition, which is termed the Dirty Model \cite{jalali2010dirty}:
\begin{equation} \label{eq:dirty}
\min_{\mathbf{A},\mathbf{B}}\frac{1}{2}||\mathbf{X} - \mathbf{D} (\mathbf{A} + \mathbf{B})||_F^2 + \lambda_1 ||\mathbf{A}||_{1,\infty} + \lambda_2 ||\mathbf{B}||_{1,1},
\end{equation}
where $||\cdot||_F$ denotes the Frobenius norm, $l_{1,\infty}$-norm encourages the block sparsity and $l_{1,1}$-norm promotes sparsity. The Dirty model addresses the drawback of C-HiLasso because $\mathbf{A}$ points out dictionary atoms that are shared across all tasks (similarity) and $\mathbf{B}$ captures those that are uniquely utilized by individual task (difference). However, it assumes no label differences between dictionary atoms, thus it lacks the group information that indicates sub-dictionaries for different classes.

In summary, there are three key factors one could consider when designing DL methods for classification: sparsity, group structure and if possible, within-group similarity. Sparsity makes it easier to interpret the data and brings in the possibility of identifying the difference in a high-dimensional feature space. Group structure naturally coincides with the label information in the classification problem. It enforces the labels implicitly, thus will not increase the size of the problem. Within-group similarity can be used to further refine the group structure by finding a smaller set of dictionary atoms in each group that can resemble all the data in each class.

Inspired by this observation, we propose the framework of structured dictionary learning $-$ StructDL with a single task version, Hierarchical Dictionary Learning (HiDL) and a multi-task version, Group Structured Dirty Dictionary Learning (GDDL) as in Fig \ref{fig:StructDL}. Different from sparsity or locality driven DL approaches, HiDL strictly enforces the group boundary between different classes, thus works better when the data is close to the class boundary. As an extension of HiDL to multi-task scenario, GDDL combines the group structure with the Dirty Model so that we could find the shared atoms from in each class. This could further strength the locality within each group since the shared dictionary atoms will be more compact in a small neighborhood as in Fig \ref{fig:StructDL}(d). Notice that constraint functions $f_{\mathbf{A}}(\cdot)$ and $f_{\mathbf{D}}(\cdot)$ mentioned in Section I.B could also be merged into the StructDL framework. However, we adhere to a simple formulation to better understand the principles that matter in following sections.

\subsection{Hierarchical Dictionary Learning (HiDL)}
When training data has large within-class variability, it makes more sense to utilize sparse coding in a single task setup than leveraging correlation in multi-task coding. A properly structured mapping enforced by HiLasso (\ref{eq:hilasso}) in DL process can guarantee that dictionary atoms are only updated by training data from same class. This implicit label consistency between dictionary atoms and data can not be enforced by either Lasso or LLC. Thus, we propose the single task version of StructDL $-$ Hierarchical Dictionary Learning (HiDL), whose objective function is
\begin{equation} \label{eq:HiDL}
\min_{\mathbf{D},\mathbf{A}}\sum_{i=1}^{N}{(\frac{1}{2}||\mathbf{x}_i - \mathbf{D} \mathbf{a}_i||_2^2 + \lambda_1 \sum_{g \in \mathcal{G}}^{}{||\mathbf{a}_{i,[g]}||_2}} + \lambda_2 ||\mathbf{a}_i||_1),
\end{equation}
essentially incorporating HiLasso into DL process. Similar to other DL methods, HiDL iterates between sparse coding and dictionary update. For the sparse coding stage, we are solving HiLasso problem with a well-defined group structure. Convex optimization based approaches \cite{sprechmann2011c,bach2012structured} or Bayesian approach using structured Spike and Slab prior \cite{suo2013hierarchical} can be adopted for this purpose. 

For the dictionary update stage, we adopt the method of block coordinate descent with a warm start to update one dictionary atom at a time \cite{mairal2009online}. Furthermore, we will show in Section III that under certain conditions this approach forces the dictionary atoms to be updated in the same subspace. Using the facts that $||\mathbf{X} - \mathbf{D}\mathbf{A}||_F^2 = tr[(\mathbf{X} - \mathbf{D}\mathbf{A})(\mathbf{X} - \mathbf{D}\mathbf{A})^\top] $ and trace is invariant under cyclic permutations, the objective function of the dictionary update step can be changed to:
\begin{equation} 
\min_{\mathbf{D}}\frac{1}{2}tr(\mathbf{D}^\top\mathbf{D}\mathbf{\Psi}) - tr(\mathbf{D}^\top\mathbf{\Phi})
\end{equation}
where 
\begin{equation} \label{eq:dictUpdate1}
\mathbf{\Psi} = [\bm{\psi}_1,...,\bm{\psi}_K]=\sum_{i=1}^{N}{\mathbf{a}_i\mathbf{a}_i^\top}
\end{equation}
and
\begin{equation} \label{eq:dictUpdate2}
\mathbf{\Phi} = [\bm{\phi}_1,...,\bm{\phi}_K]=\sum_{i=1}^{N}{\mathbf{x}_i\mathbf{a}_i^\top}.
\end{equation}
Taking the derivative and set it to zero, we obtain the dictionary update procedure as follow:
\begin{equation} \label{eq:dictUpdate3}
\hat{\mathbf{d}} \leftarrow \frac{1}{\Psi_{j,j}}(\bm{\phi}_j - \mathbf{D}\bm{\psi}_j) + \mathbf{d}_{j}^{t}\
\end{equation}
and
\begin{equation} \label{eq:dictUpdate4}
\mathbf{d}_{j}^{t+1} \leftarrow \frac{1}{max(||\hat{\mathbf{d}}||_2,1)}\hat{\mathbf{d}}
\end{equation}
where $\Psi_{j,j}$ is the value of $\mathbf{\Psi}$ at coordinate $[j,j]$ with $\mathbf{d}_{j}^{t}$ and $\mathbf{d}_{j}^{t+1}$ being the $j$-th atom at $t$-th and $t+1$-th iterations, respectively. According to (II.10), dictionary atoms always have unit norm.

Putting together the sparse coding and dictionary update processes, we complete the algorithm for StructDL as presented in Algorithm~\ref{algo:StructDL}. The dictionary is initialized with random sampling of training data and the motivation will be explained in Section III from a theoretical standpoint.

\begin{algorithm}
\DontPrintSemicolon 
\KwIn{Labeled training data $\mathbf{x}_i, i=1,..,N$, the group structure $\mathcal{G}$, scalar $\rho = 1.1$, and regularization parameters $\lambda_1$ and $\lambda_2$;}

\KwOut{Dictionary $\mathbf{D}$ and sparse code $\mathbf{A}$ (and $\mathbf{B}$);}

Initializing $\mathbf{D}^0$ by random sampling from training data of each class and $t = 0$; \;
\While{not converged}{
Fix $\mathbf{D}^{t}$ and update $\mathbf{A}^{t+1}$ using convex optimization to solve HiLasso \cite{sprechmann2011c} or Algorithm 2 to solve Group Structured Dirty Model problem.
\\
Fix $\mathbf{A}^{t+1}$ and update $\mathbf{D}^{t+1}$ using (II.7)-(II.10).
\\
Increment $t$.
}
\Return{dictionary $\mathbf{D}$ and sparse code $\mathbf{A}$ (and $\mathbf{B}$).}
\caption{Structured Dictionary Learning (StructDL)}
\label{algo:StructDL}
\end{algorithm}

\begin{figure*}
\centering
\subfigure[Dirty Model]{\includegraphics[height=1.8in]{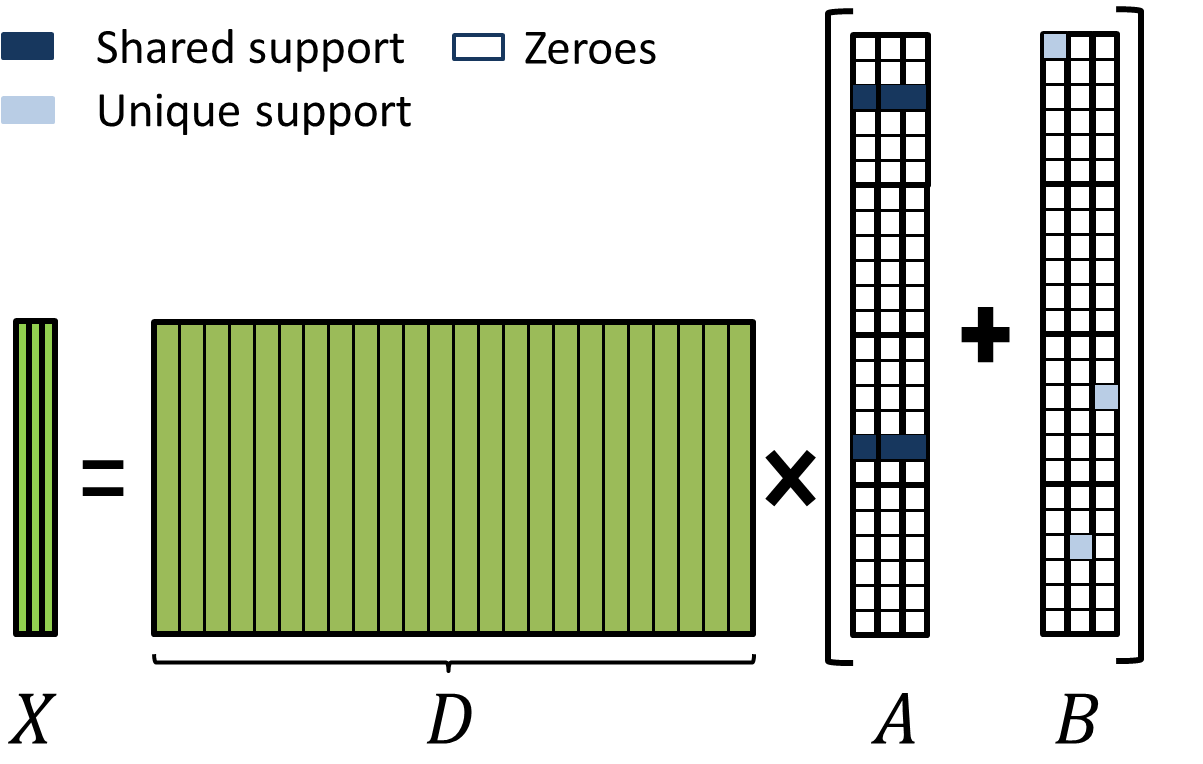}}
\subfigure[Group Structured Dirty Model]{\includegraphics[height=1.8in]{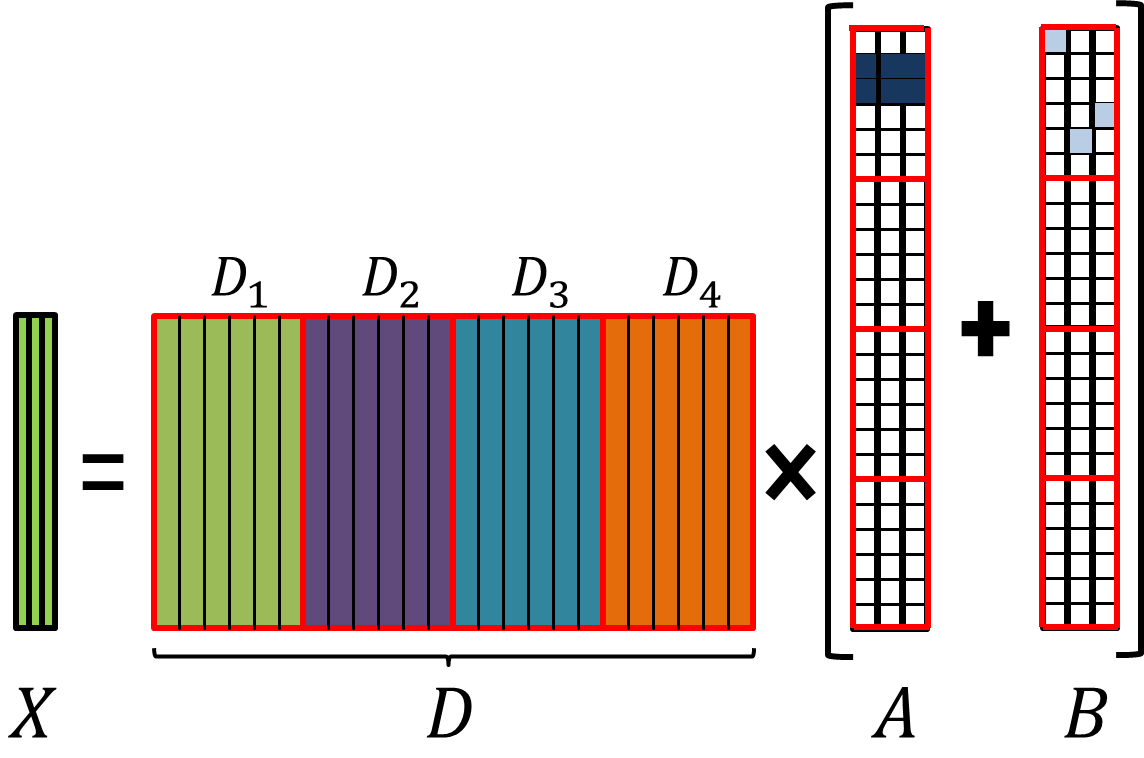}}
\caption{Comparison between the signal models of the Dirty Model and GDDL. Data $\mathbf{X}$ belongs to the same class. For the Dirty Model, the dictionary $\mathbf{D}$ only contains atoms for the same class while that of GDDL contains sub-dictionaries for four different classes, i.e., $\mathbf{D}_1,..., \mathbf{D}_4$. The sparse coefficients $\mathbf{A}$ and $\mathbf{B}$ for GDDL are forced to capture the shared supports (dark blue) and unique supports (light blue) within the group boundary (red line), while the Dirty Model does not impose such constraint.}
\label{fig:GDDL}
\end{figure*}

\subsection{Group Structured Dirty Dictionary Learning (GDDL)}
HiDL makes the assumption that different tasks are independent on how they select dictionary atoms, therefore the sparse coding step for each task is carried out separately. In some applications, training data in each class is tightly clustered, indicating a large within-class similarity. For instance, pictures of the same person taken under different illumination conditions in face recognition tasks can still be visually identified to belong to same class. Such correlation among training data with the same label is not properly captured by HiDL. Therefore, we propose a multi-task extension of HiDL $-$ Group Structured Dirty Model Dictionary Learning (GDDL) as below:
\begin{flalign} \label{eq:GDDL}
\min_{\mathbf{D},\mathbf{A}, \mathbf{B}} & \frac{1}{2}||\mathbf{X}_c - \mathbf{D} (\mathbf{A}_c + \mathbf{B}_c)||_F^2 + \lambda_1||\mathbf{A}_{c}||_{1,2} + \lambda_2||\mathbf{B}_{c}||_{1,1} \nonumber
\\
 & + \lambda_3\sum_{g \in \mathcal{G}}^{}||\mathbf{A}_{c,[g]}||_{F} + \lambda_4\sum_{g \in \mathcal{G}}^{}||\mathbf{B}_{c,[g]}||_{F}
, \forall \;c,
\end{flalign}
where $\mathbf{X}_c$ is all training data from $c$-th class, while $\mathbf{A}_{c}$ and $\mathbf{B}_{c}$ are the sub-matrices in $\mathbf{A}$ and $\mathbf{B}$ consisting of columns for class $c$, respectively. Furthermore, $\mathbf{A}_{c,[g]}$ and $\mathbf{B}_{c,[g]}$ are the sub-matrices by extracting rows with indices in group $g$ from $\mathbf{A}_{c}$ and $\mathbf{B}_{c}$, respectively. The first three terms impose the Dirty Model with $l_{1,2}$-norm and $l_{1,1}$-norm for promoting row sparsity and sparsity, respectively. Since the dictionary $\mathbf{D}$ contains sub-dictionaries from all classes, extra constraints are needed to guarantee the active rows from $\mathbf{A}_c$ and active indices from $\mathbf{B}_c$ fall into the same group, respectively. Inspired by C-HiLasso, we use the collaborative Group Lasso regularizers $\sum_{g \in \mathcal{G}}^{}||\mathbf{A}_{c,[g]}||_{F}$ and $\sum_{g \in \mathcal{G}}^{}||\mathbf{B}_{c,[g]}||_{F}$ to force the group boundary.

The underlying model of GDDL can be interpreted as a generalization of C-HiLasso and the Dirty Model. When different tasks do not have to share atoms, the sparse coding step of (II.11) turns into
\begin{align}
\min_{\mathbf{B}}\frac{1}{2}||\mathbf{X}_c - \mathbf{D} \mathbf{B}_c||_F^2 + \lambda_2||\mathbf{B}_{c}||_{1,1} + \lambda_4\sum_{g \in \mathcal{G}}^{}||\mathbf{B}_{c,[g]}||_{F}, \forall \;c,
\end{align}
which is exactly C-HiLasso enforcing both group sparsity and within-group sparsity. When there is no label difference between dictionary atoms (no group structure), the sparse coding step of (II.11) becomes
\begin{align} 
\min_{\mathbf{A}, \mathbf{B}}\frac{1}{2}||\mathbf{X}_c - \mathbf{D} (\mathbf{A}_c + \mathbf{B}_c)||_F^2 + \lambda_1||\mathbf{A}_{c}||_{1,2} + \lambda_2||\mathbf{B}_{c}||_{1,1}, \forall \;c,
\end{align}
which is the Dirty Model with decomposition of row sparsity and sparsity terms. 

Nevertheless, there are two key differences between GDDL and the Dirty model. First, GDDL extends the Dirty model by adding another layer of group sparsity, which is illustrated in Fig~\ref{fig:GDDL}. Different from the Dirty Model, GDDL enforces all the activate supports to stay within the same group corresponding to the desired class. Within the group, the sparse codes are further decomposed into two parts, one with supports shared across tasks and one with unique supports associated with different tasks. And the shared dictionary atoms captures the similarity among tasks. Second, the Dirty Model is oriented from a reconstruction perspective while the GDDL brings in the group structure for labeling purposes thus being geared towards classification. In short, GDDL could uniquely combine sparsity, group structure and within-group similarity (or locality) in a single formulation.

\textbf{Optimization Approach:}
The sparse coding step of GDDL $-$ the Group Structured Dirty Model problem can be reformulated as follows:
\begin{flalign}
\min_{\mathbf{A}, \mathbf{B}} & ||\mathbf{A}_{c}||_{1,2} + \lambda_2 ||\mathbf{B}_{c}||_{1,1} + \sum_{g \in \mathcal{G}}^{}(\lambda_3||\mathbf{A}_{c,[g]}||_{F} + \lambda_4||\mathbf{B}_{c,[g]}||_{F}) \nonumber 
\\
& s.t. \; \mathbf{X}_c - \mathbf{D} (\mathbf{A}_c + \mathbf{B}_c) = \mathbf{0}, \forall \; c,
\end{flalign}
with the re-scaled regularization parameters (which will not affect the results). We choose the alternating direction method of multipliers (ADMM) as the optimization approach because of its simplicity, efficiency and robustness \cite{boyd2011distributed,yang2011alternating}. By introducing two auxiliary variables $\mathbf{U} \in \mathbb{R}^{K \times N}$ and $\mathbf{V} \in \mathbb{R}^{K \times N}$, this problem can be reformulated as:
\begin{flalign}
\min_{\mathbf{A}, \mathbf{B}, \mathbf{U}, \mathbf{V}} & ||\mathbf{U}_{c}||_{1,2} + \lambda_2 ||\mathbf{V}_{c}||_{1,1} \nonumber
\\
+ &\sum_{g \in \mathcal{G}}^{}(\lambda_3||\mathbf{U}_{c,[g]}||_{F} + \lambda_4||\mathbf{V}_{c,[g]}||_{F}) \nonumber
\\
s.t. & \; \mathbf{A}_c - \mathbf{U}_c = \mathbf{0}, \; \mathbf{B}_c - \mathbf{V}_c = \mathbf{0}, \;\nonumber
\\
& \mathbf{X}_c - \mathbf{D} (\mathbf{A}_c + \mathbf{B}_c) = \mathbf{0}, \forall \; c.
\end{flalign}
Therefore, the augmented Lagrangian function with respect to $\mathbf{A}$, $\mathbf{B}$, $\mathbf{U}$, and $\mathbf{V}$ can be formed as:
\begin{flalign} \label{Lag_Dirty}
L_{\mu}(\mathbf{A},\mathbf{B},\mathbf{U}, \mathbf{V}) = 
& \sum_{c=1}^{C}\Big(||\mathbf{U}_{c}||_{1,2} + \lambda_2 ||\mathbf{V}_{c}||_{1,1} \nonumber
\\
+ & \lambda_3\sum_{g \in \mathcal{G}}^{}||\mathbf{U}_{c,[g]}||_{F} + \lambda_4\sum_{g \in \mathcal{G}}^{}||\mathbf{V}_{c,[g]}||_{F} \Big) \nonumber
\\
+ & tr(\hat{\mathbf{Y}}_{1},\mathbf{A} - \mathbf{U}) + tr(\hat{\mathbf{Y}}_{2}, \mathbf{B} - \mathbf{V}) \nonumber
\\
+ & tr(\hat{\mathbf{Y}}_{3}, \mathbf{X} - \mathbf{D} (\mathbf{A} + \mathbf{B})) \nonumber
\\
+ & \frac{\mu}{2}\Big(||\mathbf{A} - \mathbf{U}||_F^2 + ||\mathbf{B} - \mathbf{V}||_F^2  \nonumber
\\
+ & ||\mathbf{X} - \mathbf{D} (\mathbf{A} + \mathbf{B})||_F^2 \Big)
\end{flalign}
where $\hat{\mathbf{Y}}_{1}$, $\hat{\mathbf{Y}}_{2}$, $\hat{\mathbf{Y}}_{3}$ are the Lagrangian multipliers for equality constraints and $\mu > 0$ is a penalty parameter. The augmented Lagrangian function (\ref{Lag_Dirty}) can be minimized over $\mathbf{A}$, $\mathbf{B}$, $\mathbf{U}$, and $\mathbf{V}$ iteratively by fixing one variable at a time and updating the others. The entire algorithm is summarized in Algorithm~\ref{algo:GroupDirty}, where we let $\mathbf{Y}_{1} = \frac{\hat{\mathbf{Y}}_{1}}{\mu}$, $\mathbf{Y}_{2} = \frac{\hat{\mathbf{Y}}_{2}}{\mu}$, $\mathbf{Y}_{3} = \frac{\hat{\mathbf{Y}}_{3}}{\mu}$. And $\mathbf{Y}_{1,c}$, $\mathbf{Y}_{2,c}$ and $\mathbf{Y}_{3,c}$ are the submatrices with columns corresponding to $c$-th class in $\mathbf{Y}_{1}$, $\mathbf{Y}_{2}$ and $\mathbf{Y}_{3}$, respectively.

\begin{algorithm}
\DontPrintSemicolon 
\KwIn{Training data $\mathbf{X}$, learned dictionary $\mathbf{D}$, group structure $\mathcal{G}$, scalar $\rho = 1.1$, and regularization parameters $\lambda_2$,$\lambda_3$,$\lambda_4$;}

\KwOut{Sparse codes $\mathbf{A}$ and $\mathbf{B}$;}

Initializing $\mathbf{A}^0 = \mathbf{0}$, $\mathbf{B}^0 = \mathbf{0}$, $\mathbf{Y}_{1}^0= \mathbf{0}$, $\mathbf{Y}_{2}^0= \mathbf{0}$, $\mathbf{Y}_{3}^0= \mathbf{0}$, $\mu=1$, $\mu_{max} = 10^6$, $k = 0$; \;
\For{$c = 1,...,C$}{
\While{not converged}{
Fix $\mathbf{A}_c$, $\mathbf{B}_c$, $\mathbf{V}_c$ and update $\mathbf{U}_c$ by:   
\begin{align*}
\mathbf{U}_c^{k+1} & = \argmin L_{\mu}(\mathbf{A}_c^k,\mathbf{B}_c^k,\mathbf{U}_c,\mathbf{V}_c^{k}) \nonumber
\\
& = \mbox{Prox}_{\Omega_{\mathcal{G},(1,2)}}(\mathbf{A}_c^k + \mathbf{Y}_{1,c}^k) \nonumber
\end{align*}

Fix $\mathbf{B}_c$, $\mathbf{U}_c$, $\mathbf{V}_c$ and update $\mathbf{A}_c$ by:   
\begin{align*}
\mathbf{A}_c^{k+1} & = \argmin L_{\mu}(\mathbf{A}_c,\mathbf{B}_c^k,\mathbf{U}_c^{k+1},\mathbf{V}_c^k) \nonumber
\\
& = (\mathbf{D}^{\top}\mathbf{D} + \mathbf{I})^{-1} \nonumber
\\
& [\mathbf{D}^{\top}(\mathbf{X}_c +\mathbf{Y}_{3,c}^k - \mathbf{D}\mathbf{B}_c^k)+ \mathbf{U}_c^{k+1} - \mathbf{Y}_{1,c}^k] \nonumber
\end{align*}

Fix $\mathbf{A}_c$, $\mathbf{B}_c$, $\mathbf{U}_c$ and update $\mathbf{V}_c$ by:   
\begin{align*}
\mathbf{V}_c^{k+1} & = \argmin L_{\mu}(\mathbf{A}_c^{k+1},\mathbf{B}_c^k,\mathbf{U}_c^{k+1},\mathbf{V}_c) \nonumber
\\
& = \mbox{Prox}_{\Omega_{\mathcal{G},(1,1)}}(\mathbf{B}_c^k + \mathbf{Y}_{2,c}^k) \nonumber
\end{align*}

Fix $\mathbf{A}_c$, $\mathbf{U}_c$, $\mathbf{V}_c$ and update $\mathbf{B}_c$ by:   
\begin{align*}
\mathbf{B}_c^{k+1} & = \argmin L_{\mu}(\mathbf{A}_c^{k+1},\mathbf{B}_c,\mathbf{U}_c^{k+1},\mathbf{V}_c^{k+1}) \nonumber
\\
& = (\mathbf{D}^{\top}\mathbf{D} + \mathbf{I})^{-1} \nonumber
\\
& [\mathbf{D}^{\top}(\mathbf{X}_c + \mathbf{Y}_{3,c}^k - \mathbf{D}\mathbf{A}_c^{k+1})+ \mathbf{V}_c^{k+1} - \mathbf{Y}_{2,c}^k] \nonumber
\end{align*}

Update Lagrange multipliers $\mathbf{Y}_{1,c}$, $\mathbf{Y}_{2,c}$, $\mathbf{Y}_{3,c}$: 
\begin{align*}
\mathbf{Y}_{1,c}^{k+1} & = \mathbf{Y}_{1,c}^{k} + \mathbf{A}_c^{k+1} - \mathbf{U}_c^{k+1} \nonumber
\\
\mathbf{Y}_{2,c}^{k+1} & = \mathbf{Y}_{2,c}^{k} + \mathbf{B}_c^{k+1} - \mathbf{V}_c^{k+1} \nonumber
\\
\mathbf{Y}_{3,c}^{k+1} & = \mathbf{Y}_{3,c}^{k} + \mathbf{X}_c - \mathbf{D}(\mathbf{A}_c^{k+1} + \mathbf{B}_c^{k+1}) \nonumber
\end{align*}

Update penalty parameter $\mu = \min(\mu_{max}, \rho\mu)$

Increment $k$.
}
}

\Return{Estimated sparse codes $\mathbf{A}$ and $\mathbf{B}$.}
\caption{Solving Group Structured Dirty Model Problem with ADMM}
\label{algo:GroupDirty}
\end{algorithm}

The key steps in Algorithm~\ref{algo:GroupDirty} are Step 4 and 6. Because Group Structured Dirty Model could be regarded as an extension of C-HiLasso as pointed out by (II.12), $\mbox{Prox}_{\Omega_{\mathcal{G},(1,1)}}$ in Step 6 can be solved using the same operator for C-HiLasso ((III.14), \cite{sprechmann2011c}), which is derived using SpaRSA framework \cite{wright2009sparse}. Although similar procedure can be carried out for Step 4 using the same framework, we follow a more straightforward approach to derive the corresponding operator. 

As pointed out in \cite{bach2011optimization}, the proximal operators associated with the composite norm in hierarchical sparse coding can be obtained by the composition of the proximal operators as long as the sparsity structures follows the right order. This order is termed as a total order relationship or tree-structured sets of groups (Definition 1, \cite{jenatton2010proximal}), which requires that the two groups are either disjoint or one is included in the other. In our case, the Group Structured Dirty Model contains group sparsity structure and row sparsity structure for $\mathbf{A}_c$ and it contains group sparsity structure and element-wise sparsity structure for $\mathbf{B}_c$. Both cases satisfy the total order relationship because either the individual index or the individual row is included in groups as clearly shown in Fig~\ref{fig:GDDL}(b). After establishing the total order relationship, the proximal operators for composite norm could be constructed by applying the proximal operators for smaller groups first, followed by the ones for larger groups. Therefore, the corresponding operators for Step 4 and 6 in Algorithm~\ref{algo:GroupDirty} can be derived as below:
\begin{equation}
\mbox{Prox}_{\Omega_{\mathcal{G},(1,2)}} = \mbox{Prox}_{\kappa_1,\Omega_{\mathcal{G}}} \circ \mbox{Prox}_{\kappa_2,\Omega_{1,2}}
\end{equation}
and
\begin{equation}
\mbox{Prox}_{\Omega_{\mathcal{G},(1,1)}} = \mbox{Prox}_{\kappa_3,\Omega_{\mathcal{G}}} \circ \mbox{Prox}_{\kappa_4,\Omega_{1,1}}
\end{equation}
where $\mbox{Prox}_{\kappa_1,\Omega_{\mathcal{G}}}$ and $\mbox{Prox}_{\kappa_3,\Omega_{\mathcal{G}}}$ are the proximal operators for group sparsity, whereas $\mbox{Prox}_{\kappa_2,\Omega_{1,2}}$ and $\mbox{Prox}_{\kappa_4,\Omega_{1,1}}$ promotes the selection of only a few non-zero rows and elements, respectively. So $\mbox{Prox}_{\Omega_{\mathcal{G},(1,2)}}$ for Step 4 can be readily computed by applying first the proximal operator associated with the $l_{1,2}$-norm (row-wise soft-thresholding) and then the one associated with group sparsity  $\mbox{Prox}_{\kappa_1,\Omega_{\mathcal{G}}}$. Similarly, the C-HiLasso operator $\mbox{Prox}_{\Omega_{\mathcal{G},(1,1)}}$ for Step 6 is just applying the element-wise soft-thresholding and then the group thresholding, which is same as in \cite{sprechmann2011c}. Here, we have $\kappa_1 =\frac{\lambda_3}{\mu}, \kappa_2 = \frac{1}{\mu}, \kappa_3 =\frac{\lambda_4}{\mu}, \kappa_4 =\frac{\lambda_2}{\mu}$. 

Inside each group, the proximal operator $\mbox{Prox}_{\kappa_2,\Omega_{1,2}}$ that encourages row sparsity is:
\begin{equation}
\mbox{Prox}_{\kappa_2,\Omega_{1,2}}(\mathbf{v}_{(j,:)}) = \left( 1 - \frac{\kappa_2}{||\mathbf{v}_{(j,:)}||_2} \right)_{+}\mathbf{v}_{(j,:)}
\end{equation}
where $\mathbf{v}_{(j,:)}$ is defined as $j$-th row of $\mathbf{V}$ and  $(x)_{+}:= \max(x,0)$. So it will zero out rows with $l_2$-norms below the threshold $\kappa_2$. The proximal operator $\mbox{Prox}_{\kappa_4,\Omega_{1,1}}$ for component-wise sparsity is:
\begin{equation}
\mbox{Prox}_{\kappa_4,\Omega_{1,1}}(v_{j,i}) = \left( 1 - \frac{\kappa_4}{|v_{j,i}|} \right)_{+}v_{j,i}
\end{equation}
where $v_{j,i}$ is the value of $\mathbf{V}$ at the coordinate $[j,i]$. Finally, the proximal operator for group sparsity is:
\begin{equation}
\mbox{Prox}_{\kappa_1,\Omega_{\mathcal{G}}}(\mathbf{V}_{[g]}) = \left( 1 - \frac{\kappa_1}{||\mathbf{V}_{[g]}||_F} \right)_{+}\mathbf{V}_{[g]}
\end{equation}
where $\mathbf{V}_{[g]}$ is the sub-matrix with rows indexed by group $g$. It has the effect of zeroing or keeping coefficients in the same group all together. Note that since GDDL separates the sparse code into shared indices $\mathbf{A}_c$ and unique indices $\mathbf{B}_c$, we observe rarely the group that wins the selection in $\mathbf{A}_c$ is different from the selection in $\mathbf{B}_c$. To avoid such scenario, we enforce the same group selection by always using the group selected by row-sparsity term, because it is a stronger constraint than sparsity.

\subsection{Classification approach}
For classification, we choose a linear classifier for its simplicity and the purpose of fair comparison with results of other techniques, although advanced classification techniques (i.e., SRC) could potentially lead to better performances. The linear classifier $\mathbf{W} \in \mathbb{R}^{C \times K}$ is found by:
\begin{equation} \label{eq:linearclassifier}
\mathbf{W}^\top = {(\mathbf{A}\mathbf{A}^\top + \eta\mathbf{I})}^{-1}\mathbf{A}\mathbf{L}^\top
\end{equation}
where $\mathbf{A}$ is the learned sparse codes for training data from either HiDL or GDDL. The matrix $\mathbf{L} \in \mathbb{R}^{C \times N}$ provides the label information for training data. If training data $\mathbf{x}_i$ belongs to the $c$-th class, then $L_{c,i}$ is one and all other elements in the same column are zero. The parameter $\eta$ controls the trade-off between the classification accuracy and the smoothness of the classifier. If the sparse coefficient $\mathbf{A}$ has block diagonal structure, so does the linear classifier $\mathbf{W}$. Thus, the non-zero sparse coefficients on undesired support could be zeroed out by the classifier. We will further explore the condition for $\mathbf{A}$ to have the block diagonal structure in Section III. For each test data $\mathbf{x}$, we find its sparse code by solving HiLasso or Group Structured Dirty Model problem with the learned dictionary $\mathbf{D}$, then apply the classifier $\mathbf{W}$ to get the label vector $\mathbf{l}_{est}$. The test data is then assigned to the class $c = \arg\max_{c} \mathbf{l}_{est}$. 

For GDDL, we only use the shared sparse coefficient $\mathbf{A}$ to train the classifier. This has the benefit of making the sparse coefficients more discriminative because they are mapped to the dictionary atoms that are within the center of the cluster. Therefore we could increase the between class distance among the sparse codes of different classes. For the subsequent classification step, we only feed the shared sparse code $\mathbf{a}$ into the classifier.

\section{Theoretical Analysis}
In this section, we will focus on HiDL and present theoretical guarantees to justify the benefit and tradeoff of using structured sparsity in DL for classification. Currently, most of the theoretical analysis of DL focused on the properties of the learned dictionary from a reconstruction perspective. It has been shown that given enough noiseless or small Gaussian noise contaminated training data, using $l_1$-or $l_0$-norm regularization in DL leads to a dictionary $\mathbf{D}$, which is a local minimum around the groundtruth with high probability \cite{spielman2012exact, jenatton2012local, schnass2013identificability}. However, little theoretical effort is focused on analyzing the discrimination power of the learned dictionary, which we will explore in this section.  

The DL problem is non-convex, making the direct analysis of its solution not trivial. Inspired by the connection between K-SVD and K-means, we interpret the sparse coding stage as analogous to sparse subspace clustering (SSC) \cite{elhamifar2012sparse}, and the dictionary learning step is essentially a way of learning the basis for different subspaces. However, there are two key differences between HiDL and SSC.
\newline
\textit{(i)} HiDL is proposed for classification and SSC is developed for clustering, thus the first difference is the availability of the group structure (label) information. In HiDL, different groups correspond to different subspaces (labels). This in turn leads to the enforcement of group structure sparsity rather than $l_1$-norm, which is later shown to make the condition for perfect sparse decomposition stricter. However, this price is paid to make the sparse code more discriminative by guaranteeing perfect block structure to separate different classes; 
\newline
\textit{(ii)} To represent the subspaces, HiDL uses learned dictionary atoms while SSC uses data directly. Therefore, the success of SSC only depends on the success recovery of sparse coding step since subspace representation (data) is fixed. While for HiDL, dictionary atoms are updated in every iteration so we also need to demonstrate that the dictionary update will not jeopardize the representation of the subspaces. This motivates us to take an inductive approach for analysis.

In this section, we assume that the sparse decomposition is exact so all training data have a perfect decomposition $\mathbf{x}_i = \mathbf{D}\mathbf{a}_i$. Scalings of $\lambda_1$ and $\lambda_2$ do not affect the optimal solution, so we replace them by a single parameter $\lambda$. Now the sparse coding step of HiDL could be re-written as:
\begin{equation} \label{eq:hilasso2}
\min_{\mathbf{A}}\lambda \sum_{g \in \mathcal{G}}^{}{||\mathbf{a}_{i,[g]}||_2} + (1 - \lambda) ||\mathbf{a}_i||_1 \; s.t. \; \mathbf{x}_i = \mathbf{D}\mathbf{a}_i \;, \forall i
\end{equation}
Then, we borrow the concepts of independent and disjoint subspaces from SSC framework \cite{elhamifar2012sparse} as below.

\noindent \textbf{Definition 1:}\textit{
Given a collection of subspaces $\{\mathcal{S}_c\}_{c=1}^C$. If $dim(\oplus_{c=1}^C \mathcal{S}_c) = \sum_{c=1}^C dim(\mathcal{S}_c)$, then $\{\mathcal{S}_c\}_{c=1}^C$ is independent where $\oplus$ denotes the direct sum operator. If every pair of subspaces intersect only at the origin, then $\{\mathcal{S}_c\}_{c=1}^C$ is disjoint. }

The index of subspaces ($c=1,...,C$) is purposely chosen to be same as the class labels to emphasize the correspondence between sub-dictionary $\mathbf{D}_c$ and subspace $\mathcal{S}_c$ (class label). To characterize two disjoint subspaces, \cite{elhamifar2012sparse} also defined an important notion: the smallest principal angle. 

\noindent \textbf{Definition 2:}\textit{
The smallest principle angle $\theta_{c_1,c_2}$ between two disjoint subspaces $\mathcal{S}_{c_1}$ and $\mathcal{S}_{c_2}$ is:
\begin{equation*}
cos(\theta_{c_1,c_2}) = \max_{\mathbf{v}_{c_1} \in \mathcal{S}_{c_1}, \mathbf{v}_{c_2} \in \mathcal{S}_{c_2}} \frac{\mathbf{v}_{c_1}^\top\mathbf{v}_{c_2}}{||\mathbf{v}_{c_1}||_2||\mathbf{v}_{c_2}||_2}
\end{equation*}
which gives $cos(\theta_{c_1,c_2}) \in [0,1).$}

\subsection{Performance Analysis}
With the aforementioned notations, we use an induction approach to show the following result.

\noindent \textbf{Theorem 1:}
\textit{Given enough noiseless training data points spanning all $C$ subspaces $\{\mathcal{S}_c\}_{c=1}^C$ of dimension $\{r_c\}_{c=1}^C$. If  we train the dictionary using HiDL, and both Lemma 1 (or Lemma 3) and Lemma 4 are satisfied, the noiseless test data from the same $C$ subspaces will have a perfect block sparse representation with respect to the trained dictionary.} 

To be more specific, we will show two properties that hold under certain conditions.

\noindent\textit{(i)} Support recovery property: in the sparse coding stage, the sparse code $\mathbf{a}$ for training data $\mathbf{x}$ of $c$-th class will have a perfect block structure such that $\mathbf{a}_{c} \neq 0$ and  $\mathbf{a}_{-c} = 0$, where $\mathbf{a}_{c}$ and $\mathbf{a}_{-c}$ indicate the sub-vectors corresponding to the subspace $\mathcal{S}_c$ and all other subspaces except $\mathcal{S}_c$;

\noindent\textit{(ii)} Subspace consistency property: in the dictionary learning stage, the dictionary update procedures (II.7) - (II.10) guarantee the dictionary atoms to be updated in the same subspace.

\textbf{Support recovery property:} Similar to Theorem 1 in \cite{elhamifar2012sparse}, it is straightforward to see the support recovery property holds for the case of independent subspace.

\noindent \textbf{Lemma 1:}\textit{
(Independent Subspace Case) Suppose the data are drawn from $C$ subspaces $\{\mathcal{S}_c\}_{c=1}^C$ of dimension $\{r_c\}_{c=1}^C$. Let $\mathbf{D}_c$ denotes the sub-dictionary for subspace $\mathcal{S}_c$ and $\mathbf{D}_{-c}$ denotes the sub-dictionary for all other subspaces except $\mathcal{S}_c$. Assume that every sub-dictionary $\mathbf{D}_c$ is full column rank. If these subspaces are independent, then for every input $\mathbf{x} \in \mathcal{S}_c$, (\ref{eq:hilasso2}) recovers a perfect subspace-sparse structure, i.e., the resulting solutions have $\mathbf{a}^*_{c} \neq 0$ and  $\mathbf{a}^*_{-c} = 0$.}

For the disjoint subspace case, we define $\mathbf{z}_{c_1}$ and $\mathbf{z}_{-c_1}$ as below:
\begin{flalign*}
\mathbf{z}_{c_1} = & \argmin \lambda \sum_{g \in \mathcal{G}}^{}{||\mathbf{z}_{[g]}||_2} + (1 - \lambda) ||\mathbf{z}||_1
\\
& s.t. \; \mathbf{x} = \mathbf{D}_{c_1} \mathbf{z}
\end{flalign*}
and
\begin{flalign*}
\mathbf{z}_{-c_1} = & \argmin \lambda \sum_{g \in \mathcal{G}}^{}{||\mathbf{z}_{[g]}||_2} + (1 - \lambda) ||\mathbf{z}||_1
\\
& s.t. \; \mathbf{x} = \mathbf{D}_{-c_1} \mathbf{z}.
\end{flalign*}
The support recovery property also holds for the disjoint subspace case as long as the following lemma holds.

\noindent \textbf{Lemma 2:}\textit{
(Disjoint Subspace Case) Given the same data and dictionary as in the independent subspace case above. If these subspaces are disjoint, then (\ref{eq:hilasso2}) recovers a perfect subspace sparse structure if and only if for all nonzero $\mathbf{x} \in \mathcal{S}_{c_1} \cap \oplus_{c_2 \neq c_1} \mathcal{S}_{c_2}$, 
\begin{flalign*}
& \lambda \sum_{g \in \mathcal{G}}^{}{||\mathbf{z}_{c_1,[g]}||_2} + (1 - \lambda) ||\mathbf{z}_{c_1}||_1 
\\
& < \lambda \sum_{g \in \mathcal{G}}^{}{||\mathbf{z}_{-c_1,[g]}||_2} + (1 - \lambda) ||\mathbf{z}_{-c_1}||_1.
\end{flalign*}
Note that $\mathbf{z}_{c_1,[g]}$ and $\mathbf{z}_{-c_1,[g]}$ are the sub-vectors of $\mathbf{z}_{c_1}$ and $\mathbf{z}_{-c_1}$ defined by group $g$.}

Since the condition for the disjoint subspace case in Lemma 2 does not explicitly impose the requirements on either the dictionary or the data, we further relate it to the characteristics of the data to be more intuitive, which yields the following result.

\noindent \textbf{Lemma 3:}\textit{
(Disjoint Subspace Case) Consider a collection of data points drawn from $C$ disjoint subspaces $\{\mathcal{S}_c\}_{c=1}^C$ of dimension $\{r_c\}_{c=1}^C$. If the condition
\begin{equation} \label{eq:condition}
\sigma_{min}(\mathbf{D}_{c_1}) > \frac{\left(\lambda + (1 - \lambda)\sqrt{K_{c_1}} \right)\max_{c_1 \neq c_2} cos(\theta_{c_1,c_2})}{\frac{\lambda}{\sqrt{K_{-c_1}}} + (1 - \lambda)}
\end{equation}
is satisfied, then for every nonzero input $\mathbf{x} \in \mathcal{S}_c$, (\ref{eq:hilasso2}) recovers a perfect subspace sparse structure, i.e., $\mathbf{a}_{c} \neq 0$ and  $\mathbf{a}_{-c} = 0$.}

\noindent \textbf{Proof:}

\noindent\textbf{Step 1:} First, we will find the upper bound $\beta_{c_1}$ for the left side of the original condition in Lemma 2, $\lambda \sum_{g \in \mathcal{G}}^{}{||\mathbf{z}_{c_1,[g]}||_2} + (1 - \lambda) ||\mathbf{z}_{c_1}||_1$.
Since data $\mathbf{x} \in \mathcal{S}_{c_1} \cap \oplus_{c_2 \neq c_1} \mathcal{S}_{c_2}$ and $\mathbf{D}_{c_1}$ is full column rank, we have,
\begin{equation} \label{eq:step1}
\mathbf{x} = \mathbf{D}_{c_1}\mathbf{z}_{c_1} \Rightarrow \mathbf{z}_{c_1} = (\mathbf{D}_{c_1}^\top \mathbf{D}_{c_1})^{-1}\mathbf{D}_{c_1}^\top\mathbf{x}
\end{equation} 
Since the subspace structure matches the group structure, we have
\begin{equation*}
\lambda \sum_{g \in \mathcal{G}}^{}{||\mathbf{z}_{c_1,[g]}||_2} + (1 - \lambda) ||\mathbf{z}_{c_1}||_1 = \lambda||\mathbf{z}_{c_1}||_2 + (1 - \lambda) ||\mathbf{z}_{c_1}||_1.
\end{equation*}
Applying the vector norm property yields
\begin{equation*}
\lambda||\mathbf{z}_{c_1}||_2 + (1 - \lambda) ||\mathbf{z}_{c_1}||_1 \leq  
\lambda||\mathbf{z}_{c_1}||_2 + (1 - \lambda)\sqrt{K_{c_1}}||\mathbf{z}_{c_1}||_2
\end{equation*}
where $K_{c_1}$ is the size of sub-dictionary $\mathbf{D}_{c_1}$. Next, applying (\ref{eq:step1}) and the matrix norm properties ($||\mathbf{A}\mathbf{x}||_2 \leq ||\mathbf{A}||_{2,2}||\mathbf{x}||_2$ and $||\mathbf{A}^{-1}||_{2,2} = \frac{1}{\sigma_{min}(\mathbf{A})}$) , we have
\begin{flalign*}
& \left(\lambda + (1 - \lambda)\sqrt{K_{c_1}}\right)||\mathbf{z}_c||_2
\\
& = \left(\lambda + (1 - \lambda)\sqrt{K_{c_1}}\right)||(\mathbf{D}_{c_1}^\top \mathbf{D}_{c_1})^{-1}\mathbf{D}_{c_1}^\top\mathbf{x}||_2 
\\
& \leq \left(\lambda + (1 - \lambda)\sqrt{K_{c_1}}\right)||(\mathbf{D}_{c_1}^\top \mathbf{D}_{c_1})^{-1}\mathbf{D}_{c_1}^\top||_{2,2}||\mathbf{x}||_2
\\
& = \frac{\lambda + (1 - \lambda)\sqrt{K_{c_1}}}{\sigma_{min}(\mathbf{D}_{c_1})}||\mathbf{x}||_2 = \beta_{c_1}
\end{flalign*}
where $\sigma_{min}(\mathbf{D}_{c_1})$ is the smallest singular value of $\mathbf{D}_{c_1}$. Thus, we have derived the upper bound $\beta_{c_1}$ for the left side of the condition.

\noindent\textbf{Step 2:} We will now show the lower bound $\beta_{-c_1}$ for the right side of the condition $\lambda \sum_{g \in \mathcal{G}}^{}{||\mathbf{z}_{-c_1,[g]}||_2} + (1 - \lambda) ||\mathbf{z}_{-c_1}||_1$.
Notice that we have
\begin{align*}
\lambda \sum_{g \in \mathcal{G}}^{}{||\mathbf{z}_{-c_1,[g]}||_2} + (1 - \lambda) ||\mathbf{z}_{-c_1}||_1
\\
= \lambda \sum_{c_2 \in \mathcal{G} \backslash c_1}^{}{||\mathbf{z}_{c_2}||_2} + (1 - \lambda) ||\mathbf{z}_{-c_1}||_1
\end{align*}
where we have abused the notation $c_2 \in \mathcal{G} \backslash c_1$ to mean all the groups excluding the one corresponding to the class $c_1$.
Because
\begin{equation*}
\lambda \sum_{c_2 \in \mathcal{G} \backslash c_1}^{}{||\mathbf{z}_{c_2}||_2} + (1 - \lambda) ||\mathbf{z}_{-c_1}||_1 \geq
\lambda ||\mathbf{z}_{-c_1}||_2 + (1 - \lambda) ||\mathbf{z}_{-c_1}||_1,
\end{equation*}
we can instead find the lower bound for the simplified condition $\lambda ||\mathbf{z}_{-c_1}||_2 + (1 - \lambda) ||\mathbf{z}_{-c_1}||_1$.
Based on the definition of $\mathbf{z}_{-c_1}$, we have
\begin{equation*}
||\mathbf{x}||_2^2 = \mathbf{x}^\top\mathbf{x} = \mathbf{x}^\top\mathbf{D}_{-c_1}\mathbf{z}_{-c_1}.
\end{equation*}
Using the Holder's inequalities ($|\mathbf{u}^\top\mathbf{v}| \leq ||\mathbf{u}||_{\infty}||\mathbf{v}||_1$ and $|\mathbf{u}^\top\mathbf{v}| \leq ||\mathbf{u}||_2||\mathbf{v}||_2$) , we obtain
\begin{equation*}
||\mathbf{x}||_2^2 = \mathbf{x}^\top\mathbf{D}_{-c_1}\mathbf{z}_{-c_1}\leq ||\mathbf{D}_{-c_1}^\top\mathbf{x}||_{\infty}||\mathbf{z}_{-c_1}||_1
\end{equation*}
and
\begin{equation*}
||\mathbf{x}||_2^2 = \mathbf{x}^\top\mathbf{D}_{-c_1}\mathbf{z}_{-c_1}\leq ||\mathbf{D}_{-c_1}^\top\mathbf{x}||_2||\mathbf{z}_{-c_1}||_2.
\end{equation*}
With the definition of smallest principle angle and the vector norm inequality, we can write
\begin{equation*}
||\mathbf{x}||_2^2 \leq \max_{c_2 \neq c_1} cos(\theta_{c_1,c_2})||\mathbf{D}_{-c_1}||_{max,2}||\mathbf{x}||_2||\mathbf{z}_{-c_1}||_1
\end{equation*}
and
\begin{equation*}
||\mathbf{x}||_2^2 \leq \sqrt{K_{-c_1}}\max_{c_2 \neq c_1} cos(\theta_{c_1,c_2})||\mathbf{D}_{-c_1}||_{max,2}||\mathbf{x}||_2||\mathbf{z}_{-c_1}||_2
\end{equation*}
where we use $||\mathbf{D}_{-c_1}||_{max,2}$ to denote the largest $l_2$ norm of the columns of $\mathbf{D}_{-c_1}$, which is 1 because we restrict the dictionary atoms in a convex set $\mathcal{D}$ to have unit norm.
Therefore, the lower bound for the right side can be shown to be
\begin{equation*}
\beta_{-c_1} = \frac{\lambda||\mathbf{x}||_2}{\sqrt{K_{-c_1}}\max_{c_1 \neq c_2} cos(\theta_{c_1,c_2})} + \frac{(1-\lambda)||\mathbf{x}||_2}{\max_{c_1 \neq c_2} cos(\theta_{c_1,c_2})}.
\end{equation*}

\noindent\textbf{Step 3:} Combining the lower bound in Step 2 together with the upper bound found in Step 1 gives
\begin{flalign*}
& \frac{\lambda + (1 - \lambda)\sqrt{K_{c_1}}}{\sigma_{min}(\mathbf{D}_{c_1})}||\mathbf{x}||_2
\\ 
& < \frac{\lambda||\mathbf{x}||_2}{\sqrt{K_{-c_1}}\max_{c_1 \neq c_2} cos(\theta_{c_1,c_2})} + \frac{(1-\lambda)||\mathbf{x}||_2}{\max_{c_1 \neq c_2} cos(\theta_{c_1,c_2})}
\end{flalign*}
which can be simplified to,
\begin{equation*}
\sigma_{min}(\mathbf{D}_{c_1}) > \frac{\left(\lambda + (1 - \lambda)\sqrt{K_{c_1}} \right)\max_{c_1 \neq c_2} cos(\theta_{c_1,c_2})}{\frac{\lambda}{\sqrt{K_{-c_1}}} + (1 - \lambda)}. \;\;\; \Box
\end{equation*}

\textbf{Subspace consistency property:} If the sparse coefficient $\mathbf{a}$ from the sparse coding step has the perfect block structure, the dictionary will have following property using the dictionary update procedures (II.7) - (II.10),

\noindent \textbf{Lemma 4:}\textit{
Suppose the training data $\mathbf{x}$ belongs to the $c$-th class. Assume that each sub-dictionary is full-rank. At the $t$-th iteration, if the dictionary atom $\mathbf{d}_{j}^{t-1} \in \mathcal{S}_c$ and sparse coefficient $\mathbf{a}^t$ from the previous sparse coding stage has a block structure such that $\mathbf{a}^t_{c} \neq 0$ and  $\mathbf{a}^t_{-c} = 0$, then the updated dictionary atom $\mathbf{d}_{j}^t \in \mathcal{S}_c$.}

\noindent\textbf{Proof:}

Based on the properties of subspace, it suffices to show instead that 
$\bm{\phi}_j - \mathbf{D}\bm{\psi}_j \in \mathcal{S}_c$. Notice that if $\mathbf{a}^t_{c} \neq 0$ and  $\mathbf{a}^t_{-c} = 0$, then $\mathbf{\Psi}$ will be block diagonal with block structures matching the subspace alignments. Therefore, $\bm{\psi}_{c} \neq 0$ and $\bm{\psi}_{-c} = 0$, i.e. $\mathbf{D}\bm{\psi}_j \in \mathcal{S}_c$. 

Also notice that $\mathbf{\Phi} = \sum_{i=1}^{N}{\mathbf{x}_i{\mathbf{a}^t_i}^\top} = \mathbf{D}(\sum_{i=1}^{N}{\mathbf{a}_i^*{\mathbf{a}^t_i}^\top)}$, where $\mathbf{a}_i^*$ represents the true sparse code for $\mathbf{x}_i$ that has same block structure. Therefore, $\sum_{i=1}^{N}{\mathbf{a}_i^*{\mathbf{a}^t_i}^\top}$ has the same block diagonal structure matching the subspace alignments, i.e. $\bm{\phi}_j \in \mathcal{S}_c$. Therefore, $\mathbf{d}_{j}^t \in \mathcal{S}_c$.  \hfill $\Box$

\subsection{Remark}
When $\lambda = 0$, the condition (\ref{eq:condition}) becomes
\begin{equation*}
\sigma_{min}(\mathbf{D}_{c_1}) > \sqrt{K_{c_1}}\max_{c_1 \neq c_2} cos(\theta_{c_1,c_2})
\end{equation*}
which is exactly the condition derived in Theorem 3 of \cite{elhamifar2012sparse} with the given dictionary having unit norm columns. Moreover, because $K_{c_1}$ is almost always smaller than $K_{-c_1}$, the condition for HiDL is stricter, which means that the requirement for using structured sparsity is stricter than using $l_1$-norm. This is the tradeoff paid to recover the sparse code with the right block structure in contrast to no constraints whatsoever on the support by $l_1$-norm. However, this also gives the benefit of the group structure, which is especially helpful for classification as illustrated in Fig~\ref{fig:StructDL}. Taking a closer look at the condition in (\ref{eq:condition}), on the left side, the smallest non-zero singular value of the dictionary is bounded from below, yielding a similar effect as the restricted isometry property (RIP) \cite{candes2006robust}, forcing the transformation between signal domain and coefficient domain to preserve the distance. 

The condition in (\ref{eq:condition}) also relates to the size of the dictionary such that the smaller the dictionary size (or indirectly the subspace dimension because the sub-dictionary is full rank), the more likely the condition can be satisfied. This has the benefit that when the intrinsic dimension of the signal or the dictionary size is small, HiDL is more likely to recover the perfect block structure, thus could lead to better classification performance.

In short, HiDL has been theoretically shown to be more favorable than the $l_0$- or $l_1$-norm guided DL for the task of classification for two reasons: \textit{(i)} it gives a perfect block structured sparse code at the expense of a stricter condition; and \textit{(ii)} it could lead to potentially better performance when the dictionary size or the intrinsic dimension of data is small. Note that we have assumed a noiseless condition, which will be extended to the case of Gaussian noise in future work. We have also taken an inductive approach for analysis rather than analyzing the solution of the algorithm. In next section, we will demonstrate the performance of StructDL using empirical results.

\section{Experimental Validation}
In this section, we compare the proposed structured StructDL approaches $-$ HiDL and GDDL to various existing dictionary learning methods for both synthetic and real datasets, such as face recognition and object classification. The public datasets used in this section are the Extended Yale B Face Database \cite{georghiades2001few}, the AR Face Database \cite{martinez1998ar}, and the Caltech101 Dataset \cite{fei2007learning}. The benchmark algorithms are Sparse Representation-based Classification (SRC) \cite{wright2009robust}, K-SVD \cite{aharon2006img}, Dictionary Learning with Structured Incoherence (DLSI) \cite{ramirez2010classification}, Discriminative K-SVD (D-KSVD) \cite{zhang2010discriminative}, Locality-constrained Linear Coding (LLC) \cite{wang2010locality}, Fisher Discrimination Dictionary Learning (FDDL) \cite{yang2011fisher}, and Label Consistent K-SVD (LC-KSVD) \cite{jiang2011learning}. We use classification accuracy and a concept called sparse code discrimination index (SDI) for comparison purposes. The classification accuracy is defined as the percentage of correctly classified test data.

\subsection{Parameter Selection}
\begin{figure}
\centering
\includegraphics[width=2.5in]{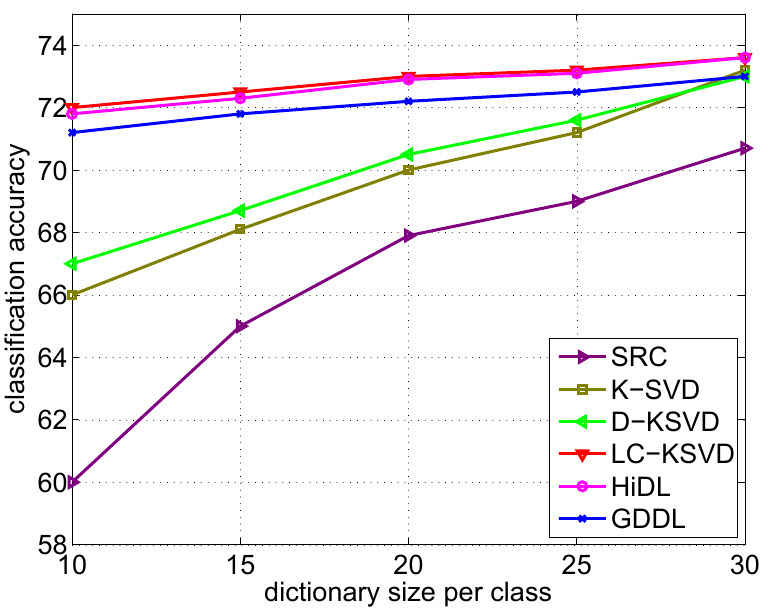}
\caption{Effect of dictionary size on classification performance of different DL methods. For Caltech 101 dataset, the size of training samples per class is fixed to 30. The dictionary atoms per class is varied from 10 to 30. As can be seen, HiDL, GDDL and LC-KSVD outperforms SRC, K-SVD and D-KSVD. GDDL does not perform as well as HiDL because of the nature of the dataset. The benefit of adding hierarchical sparsity is especially helpful when the dictionary size is small.}
\label{fig:dictsize}
\end{figure}

\textbf{Dictionary Size:} In all experiments, the initial dictionary for both HiDL and GDDL are random selections from training data with the motivation justified in Section III. As shown in \cite{yang2011fisher,jiang2011learning}, the larger the dictionary size is, the better classification performance it can generally yield. The drawback of a large dictionary size is that the size of problem becomes large simultaneously. Therefore, the ideal dictionary learning method is the one that can achieve a certain level of high performance using a small dictionary size. To compare the proposed method with other approaches on this front, we use the Caltech101 Dataset as an example. For each class, we randomly choose 30 samples for training and the rest for testing. The number of dictionary atoms for each class varies from 10 to 30. As shown in Fig~\ref{fig:dictsize}, all DL methods improve when the dictionary size becomes larger. Also, as proved in the previous section, our proposed HiDL and GDDL are comparable to LC-KSVD and all three methods consistently outperforms other sparsity driven approaches. This is consistent with our analysis in Section III.B. GDDL does not perform as well as HiDL for this dataset, probably because the dataset has very large within-class variability so the group structured dirty model does not fit the nature of the data. In contrast to other methods, HiDL and GDDL enforces label consistency implicitly using structured sparsity instead of adding extra constraint $f_{\mathbf{A}}(\cdot)$, therefore controlling the problem size.

\textbf{Regularization Parameters:} The choice of regularization parameters depends on the application and data. If a Bayesian approach is chosen for the sparse coding step, it will allow us to understand the connection between regularization parameters and data characteristics \cite{suo2013hierarchical}. Here we adopt the convex optimization based approach, thus we use cross validation to find the parameters that give the best results. 

\begin{figure}[t]
\centering
\subfigure[Overall objective function]{\includegraphics[width=1.58in]{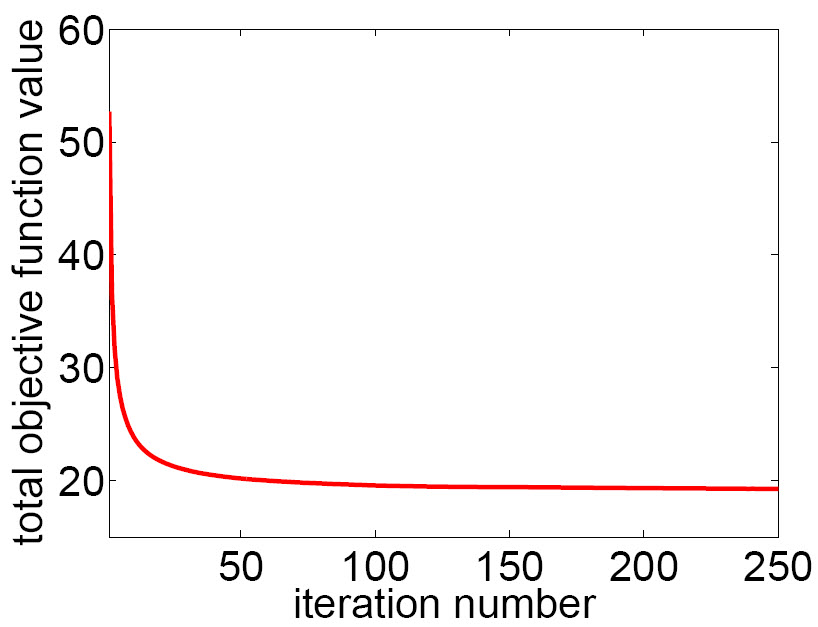}}
\subfigure[Data fidelity term]{\includegraphics[width=1.58in]{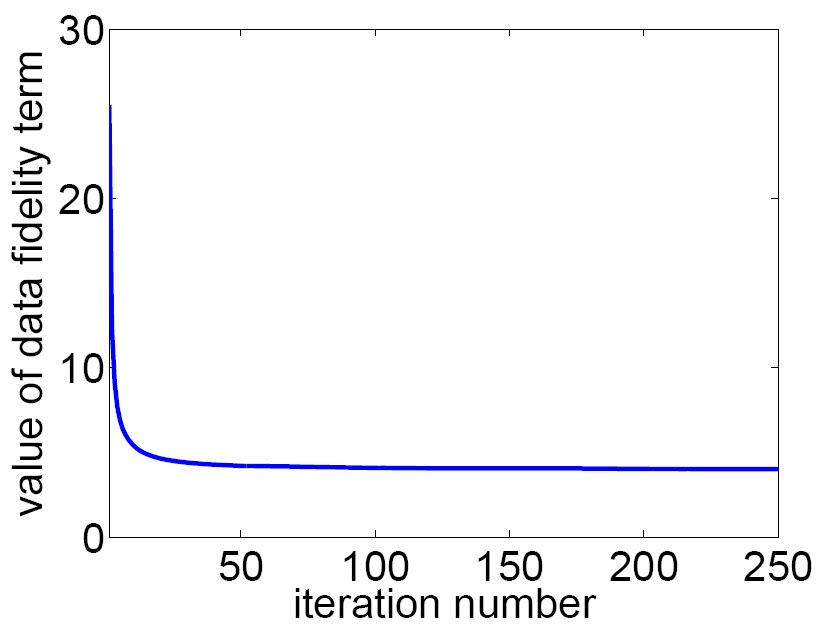}}
\subfigure[Regularization on $\mathbf{A}$]{\includegraphics[width=1.58in]{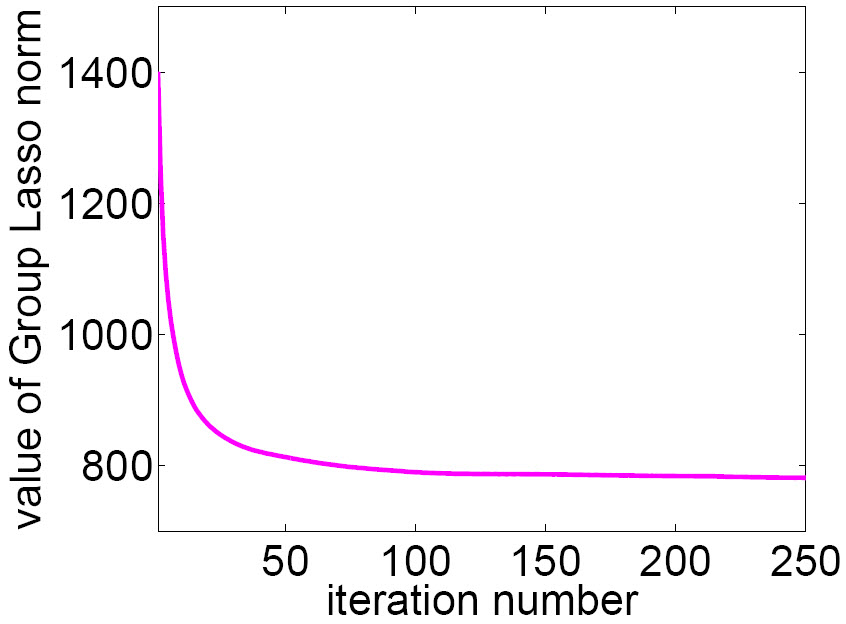}}
\subfigure[Regularization on $\mathbf{B}$]{\includegraphics[width=1.58in]{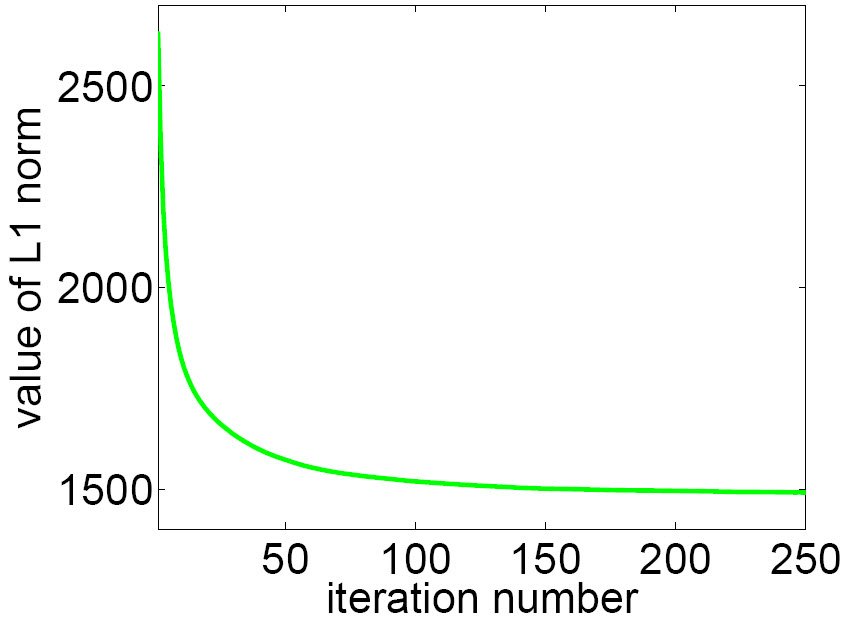}}
\caption{Convergence of GDDL using the Extended Yale B dataset. The convergence of total objective function, the data fidelity term $||\mathbf{X} - \mathbf{D} \mathbf{A} + \mathbf{B})||_F^2$, the regularization on $\mathbf{A}$ \big($\sum_{c=1}^{C}(\lambda_1||\mathbf{A}_{c}||_{1,2}+\lambda_3\sum_{g \in \mathcal{G}}^{}||\mathbf{A}_{c,[g]}||_{F})$\big) and the regularization on $\mathbf{B}$ \big($\sum_{c=1}^{C}(\lambda_2||\mathbf{B}_{c}||_{1,1}+\lambda_4\sum_{g \in \mathcal{G}}^{}||\mathbf{B}_{c,[g]}||_{F})$\big) are shown in (a), (b), (c) and (d), respectively.}
\label{fig:convergence}
\end{figure}

\begin{figure*}
\centering
\subfigure[SDI for sparsity of 5]{\includegraphics[width=2.1in]{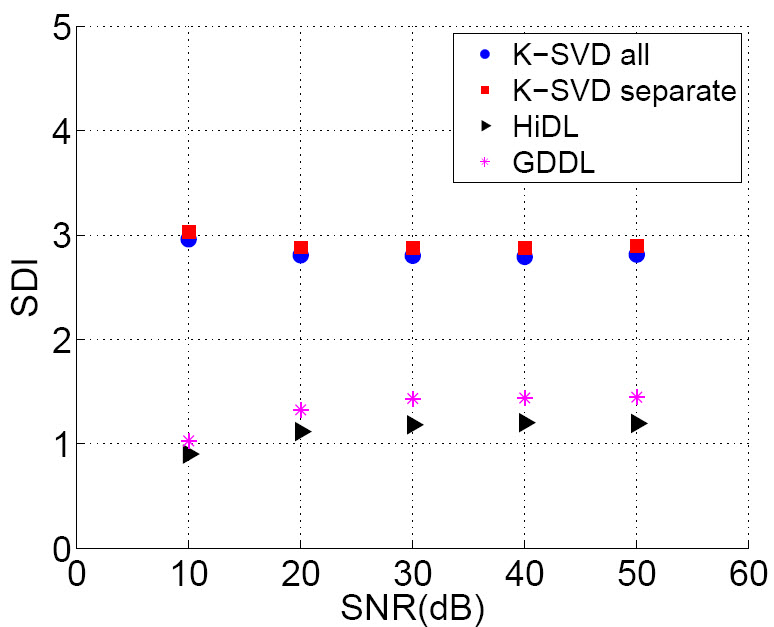}} 
\subfigure[SDI for sparsity of 25]{\includegraphics[width=2.1in]{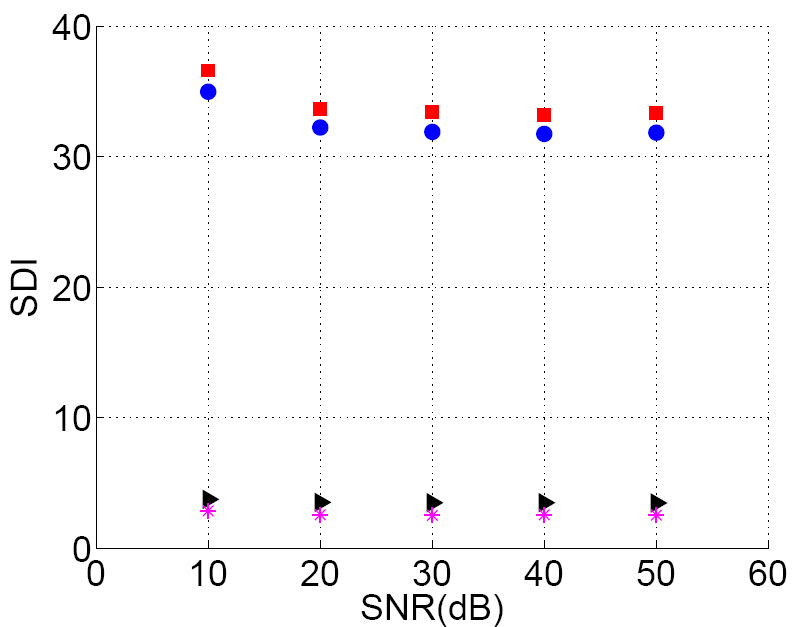}} 
\subfigure[SDI for sparsity of 40]{\includegraphics[width=2.1in]{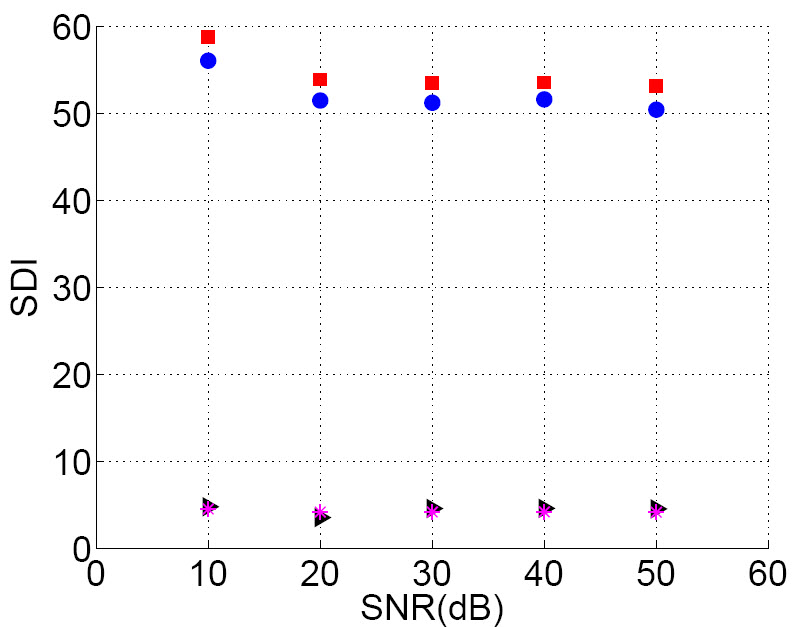}} 
\caption{Comparison of SDI using dictionaries learned from different approaches. Under different SNRs and sparsity ratios, the sparse codes generated by both HiDL and GDDL are more discriminative than either \emph{K-SVD all} or \emph{K-SVD separate}.}
\label{fig:compareHiDL}
\end{figure*}

\textbf{Stopping rule:} The stopping rule for HiDL and GDDL could be such that either the change of objective function in (\ref{eq:HiDL}) and (\ref{eq:GDDL}) are small enough or the maximum iteration number has been reached. The objective function of both HiDL and GDDL are non-convex, thus the proposed algorithm cannot find a global optimal solution. For the $l_1$-norm regularized DL \cite{mairal2009online}, it is shown that a stationary point could be found if the sufficient condition for the uniqueness of sparse coding step is satisfied. In \cite{sprechmann2011c,simon2013sparse}, the authors also prove such condition for the HiLasso norm. Following similar methodology, we could potentially show that the proposed HiDL and GDDL do converge to a stationary point. The proof itself is beyond the scope of this paper and will be presented in our future work. Here, we only show empirically the change of the objective function using Extended Yale B dataset. As shown in Fig~\ref{fig:convergence} for GDDL, the value of the whole objective function in (\ref{eq:GDDL}), the data fidelity term, the $l_{1,2}$-norm and collaborative Group Lasso norm, and the $l_{1,1}$-norm and collaborative Group Lasso norm converge around 100 iterations. The experiment setup will be described in Section IV.D. 

\subsection{Synthetic Dataset}
Unlike reconstruction-oriented dictionary learning, the StructDL framework is geared towards the task for classification. The proposed HiDL and GDDL use the group structure $\mathcal{G}$ to enforce the label consistency between sub-dictionaries and training data. Such mapping could also be realized by training a sub-dictionary $\mathbf{D}_c$ ($c = 1,...,C$) for each class independently using any previously mentioned DL methods and then concatenating the sub-dictionaries together to build $\mathbf{D} = [\mathbf{D}_1,  ..., \mathbf{D}_C]$. To understand the difference, we compare the proposed HiDL and GDDL with two different approaches: K-SVD training using data from all classes and K-SVD training for each class separately and then concatenating the dictionaries. For simplicity, we refer to them as \emph{K-SVD all} and \emph{K-SVD separate}, respectively. Note that the \emph{K-SVD separate} could be regarded as Group Lasso based DL with only one group chosen.

\textbf{Experiment Setup:} To be more specific, we would like to compare DL using the proposed structured sparsity models with DL using $l_0$-norm and Group Lasso norm under different sparsity setting and signal-to-noise ratio (SNR) levels. The true sub-dictionaries $\mathbf{D}_c$ are generated for 10 different classes ($C = 10$). Each sub-dictionary is a 20 by 50 random Gaussian matrix with unit $l_2$-norm for each column. Therefore, the group structure $\mathcal{G}$ is 10 groups with 50 sub-dictionary atoms in each group. For each class, the data $\mathbf{x}_i (i = 1,...,1500)$ is a random combination of dictionary atoms from the same sub-dictionary while the values of $\mathbf{a}_i$ are drawn from a random Gaussian distribution with zero mean and unit standard deviation. The sparsity of $\mathbf{a}_i$ are set to 5, 25 and 40 to simulate different levels of within-group sparsity. When the sparsity is 5, the within-group variation is more prominent while the within-group similarity is more significant when sparsity is 40. By concatenating data from all 10 classes, the data matrix $\mathbf{X}$ is of dimension 20 by 15000. Furthermore, zero-mean Gaussian noise is added to the data so that the SNR ranges from 10 to 50dB. Under each noise level, the experiment is repeated 10 times and each time the data is randomly splitting into two halves, training and test set.

Since there is no class label for the dictionary learned by \emph{K-SVD all},we choose the top 50 dictionary atoms corresponding to the largest coefficients for the training data in each class. The input parameters of sparsity for both \emph{K-SVD all} and \emph{K-SVD separate} are set to the true values. For HiDL and GDDL, all regularization parameters are set to 0.1, 0.05 and 0.01 for each of three sparsity levels, respectively.

\textbf{Criteria:} To measure the discriminatory power of the sparse code for both training and test data, sparse code discrimination index (SDI) is defined based on Fisher discrimination criterion \cite{yang2011fisher}:
\begin{equation} \label{eq:SDI}
SDI = \frac{1}{N}\left[ tr(\mathbf{S}_{within}(\mathbf{A})) - tr(\mathbf{S}_{between}(\mathbf{A})) \right].
\end{equation}
The with-in cluster scatter measure $\mathbf{S}_{within}(\mathbf{A})$ is defined as

\begin{equation*}
\mathbf{S}_{within}(\mathbf{A}) = \sum_{c = 1}^C \sum_{\mathbf{a}_i \in \mathbf{A}_c}(\mathbf{a}_i - \mathbf{m}_c)(\mathbf{a}_i - \mathbf{m}_c)^\top
\end{equation*}
where $\mathbf{A}_c$ is the sub-matrix formed by extracting the columns in $\mathbf{A}$ that corresponds to the $c$-th class. Here, $\mathbf{m}_c$ is the mean column vector of $\mathbf{A}_c$. The between-class scatter $\mathbf{S}_{between}(\mathbf{A})$ can be calculated by:
\begin{equation*}
\mathbf{S}_{between}(\mathbf{A}) = \sum_{c = 1}^C N_c(\mathbf{m}_c  - \mathbf{m})(\mathbf{m}_c  - \mathbf{m})^\top
\end{equation*}
where $\mathbf{m}$ is the mean column vector of $\mathbf{A}$ and $N_c$ is the number of signal in $c$-th class. A smaller SDI indicates a smaller within-class scatter and a larger between-class scatter, thus corresponding to a more discriminative sparse code. Notice that for GDDL, we only use the sparse coefficient $\mathbf{A}$ corresponding to the shared support to calculate SDI, which is also what we use for classification.

\textbf{Remark:} 
The simulation result is shown in Fig~\ref{fig:compareHiDL}. The sparse code of the test data is found with respect to the learned dictionary and the corresponding SDI is calculated using (\ref{eq:SDI}). For different in-group sparsity and SNR levels, the SDI for both HiDL and GDDL are consistently much smaller compared to that of either \emph{K-SVD all} or \emph{K-SVD separate}. Notice that the when the in-group sparsity grows, the SDI grows as well. However, the change of SDI for HiDL and GDDL does not fluctuate as much as that of both K-SVD based methods. For the case of sparsity being 25, the within-group variation and within-group similarity is balanced in some extent. Thus, GDDL works especially well when the mixture model could better suit the data as shown in Fig~\ref{fig:compareHiDL}(b), but not as well when the within-group variation or the within-group similarity is high as illustrated in Fig~\ref{fig:compareHiDL}(a) and (c), respectively. We will explore the theoretic as well as the understanding aspect of these phenomena in our future work. The results are similar for the training data and therefore omitted. 

In summary, StructDL has the advantage of forcing the sub-dictionaries for different classes to compete against each other in the sparse coding step and only the 'winners' get updated in the following dictionary update stage. Furthermore, the group structure $\mathcal{G}$ could ideally restrict the sparse codes for different classes to live in different subspaces, therefore also improving the discriminative power of the sparse codes. As pointed out in Section II.A, structured sparsity incorporating the sparsity, locality and grouping can lead to a more discriminative dictionary as HiDL and GDDL do.

\subsection{Object Classification}
\begin{figure}
\centering
\subfigure[accordion]{\includegraphics[width=3.5in]{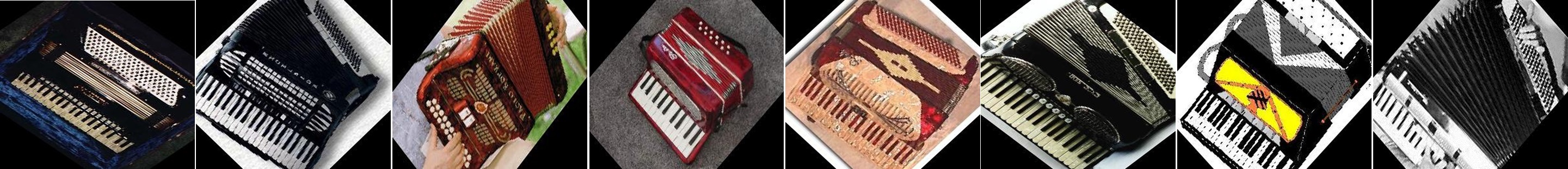}}
\subfigure[car]{\includegraphics[width=3.5in]{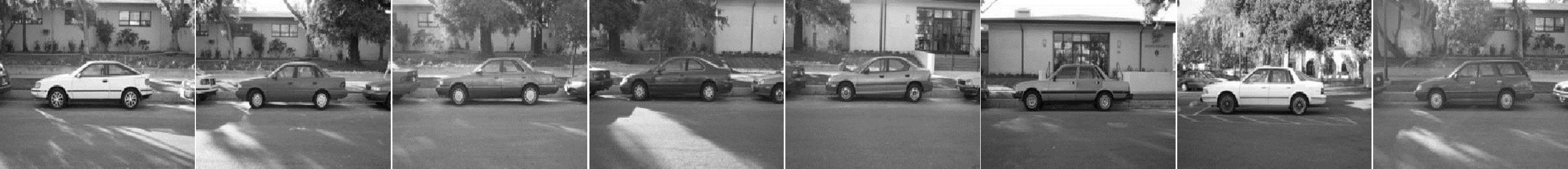}}
\subfigure[motorbikes]{\includegraphics[width=3.5in]{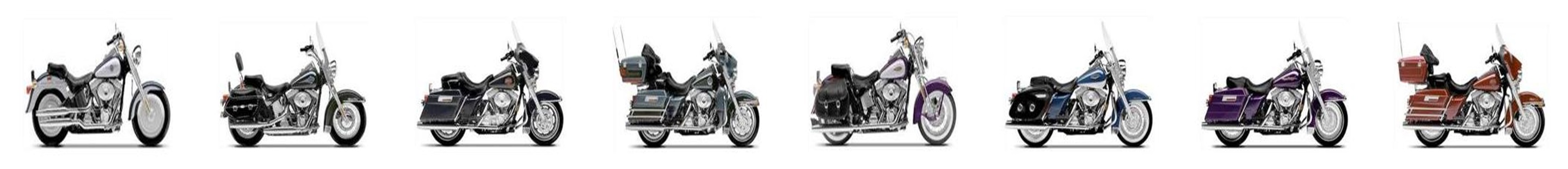}}
\subfigure[trilobite]{\includegraphics[width=3.5in]{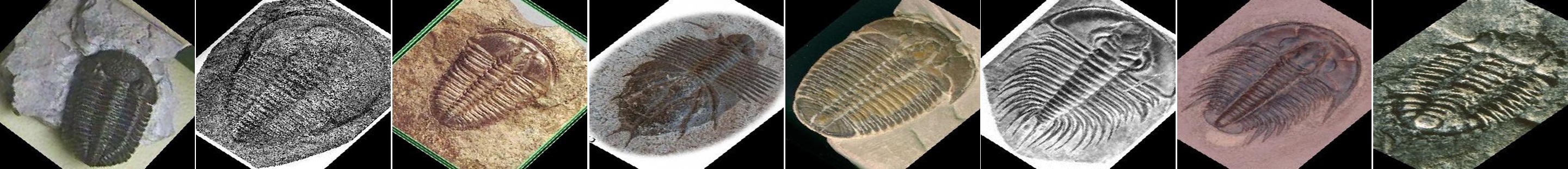}}
\caption{Examples of categories in Caltech 101 that achieve 100\% classification accuracy by HiDL.}
\label{fig:Caltech101}
\end{figure}

The Caltech 101 dataset contains 9,144 images in 102 categories, including animals, cars, planes, etc. Each category has 40 to 800 images, with most categories having around 50 images. Pictures from same class have drastic shape variability and the spatial pyramid features \cite{lazebnik2006beyond} are used as the input signal, which is same as \cite{wang2010locality,jiang2011learning}. The dimension of each feature is 3000. The size of the dictionary is the same as the number of training samples per class. We vary the number of training samples per class from 10 to 30. The experiments are repeated 10 times while HiDL and GDDL are compared with K-SVD, D-KSVD, SRC, LLC, LC-KSVD. The regularization parameters for HiDL are 0.009 and 0.007 and those for GDDL are 0.005, 0.004, 0.004 and 0.007, respectively. Our results are shown in Table~\ref{tab:Caltech101} with the results of other approaches as reported by \cite{jiang2011learning}. Our proposed HiDL consistently outperforms other approaches. As pointed out early, our proposed GDDL does not perform as well as HiDL probably because this particular dataset has large within-class variability. However, it is shown later that for face datasets, GDDL outperforms HiDL. Several of the object classes
that achieve 100\% accuracy by HiDL are shown in Fig~\ref{fig:Caltech101}.

\begin{table}
\renewcommand{\arraystretch}{1}
\caption{Comparison of proposed HiDL and GDDL and other state-of-art DL methods using Caltech 101 dataset. The dictionary size of each class is the same as the training samples per class. The best results are achieved by HiDL and bolded.}
\begin{center}
\begin{tabular}{|c||c|c|c|c|c|}
\hline
Training data size per class & 10 & 15 & 20 & 25 & 30 \\
\hline
\hline
K-SVD & 59.8 & 65.2  & 68.7 & 71.0 & 73.2\\
\hline
D-KSVD & 59.5  & 65.1 &68.6  &71.1  &73.0 \\
\hline
SRC  & 60.1 &64.9  &67.7  &69.2  &70.7 \\
\hline
LLC & 59.77 & 65.43 & 67.74 &70.16  & 73.44\\
\hline
LC-KSVD & 63.1 &67.7  &70.5  &72.3  & \textbf{73.6} \\
\hline
HiDL & \textbf{63.4} & \textbf{68.1}  & \textbf{70.9}  & \textbf{72.7}  & \textbf{73.6}  \\
\hline
GDDL & 62.1 & 66.3 & 69.0  & 71.0  & 73.1  \\
\hline
\end{tabular}
\end{center}
\label{tab:Caltech101}
\end{table}

\subsection{Face Recognition}

Face recognition is an important category of image classification tasks with applications in video surveillance and mobile imaging. The two most widely used face recognition dataset are Extended Yale B database and AR databse. Captured under various lighting conditions, the Extended Yale B database consists of 2,414 frontal-face images for 38 individuals (around 64 images per person). Similarly, the AR database has over 4,000 frontal-face images for 126 individuals, which are also taken under different conditions, including facial expressions, lighting conditions, and occlusions. Same as \cite{wright2009robust,jiang2011learning}, we crop the Extended Yale B images to the dimension of 192 $\times$ 168 pixels, normalized and projected to a vector of dimension 504 using random Gaussian projection. The AR dataset is cropped to the dimension of 165 $\times$ 120 pixels, normalized and projected to a vector of dimension 540 using random Gaussian projection. For Extended Yale B, we randomly select half of the images for training and the other half for testing in each class. For each class in the AR dataset, twenty images and six images are randomly selected for training and testing, respectively. The dictionary size for Extended Yale B and AR dataset is 15 and 5 dictionary atoms for each class, respectively. Therefore, the total dictionary contains 570 and 500 atoms. The experiment is carried out 10 times with different randomly chosen partitions. The regularization parameters for HiDL are 0.01 and 0.005 and the regularization parameters for GDDL are 0.01, 0.009, 0.005 and 0.006, respectively. The average classification accuracy is again compared with D-KSVD, LLC and LC-KSVD and shown in Table~\ref{tab:YALEBandAR}. The performances of benchmark algorithms are as reported by \cite{jiang2011learning}, which have been tuned to achieve the best results. 

\begin{figure*}
\centering
\subfigure[K-SVD result (Person 1)]{\includegraphics[width=1.5in]{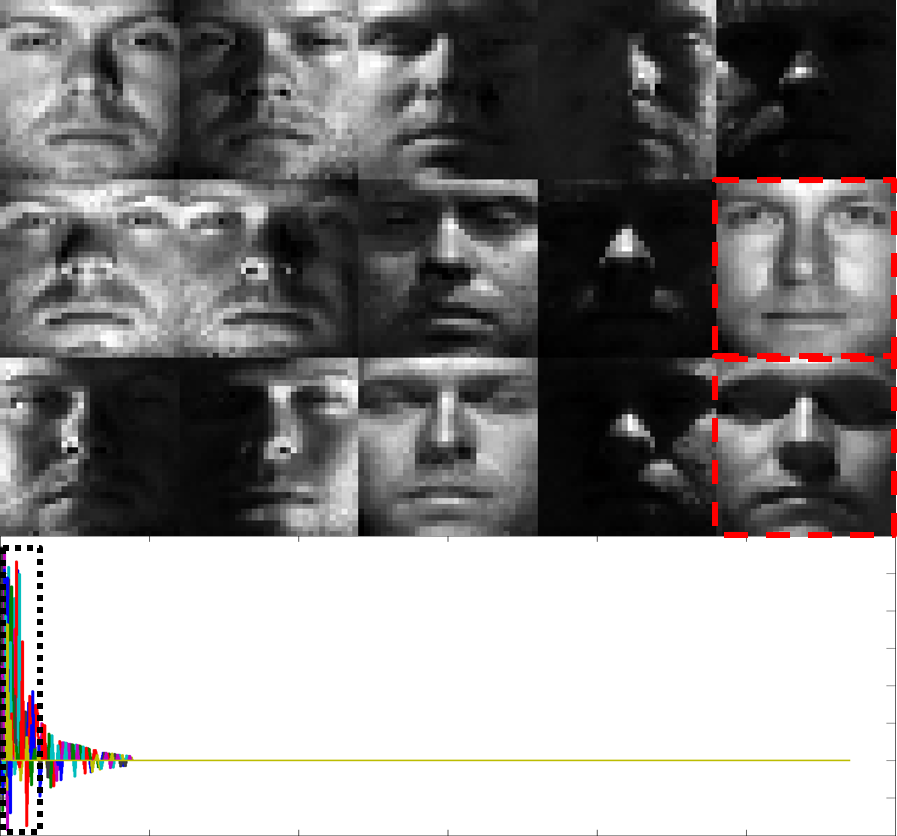}}
\subfigure[GDDL result (Person 1)]{\includegraphics[width=1.5in]{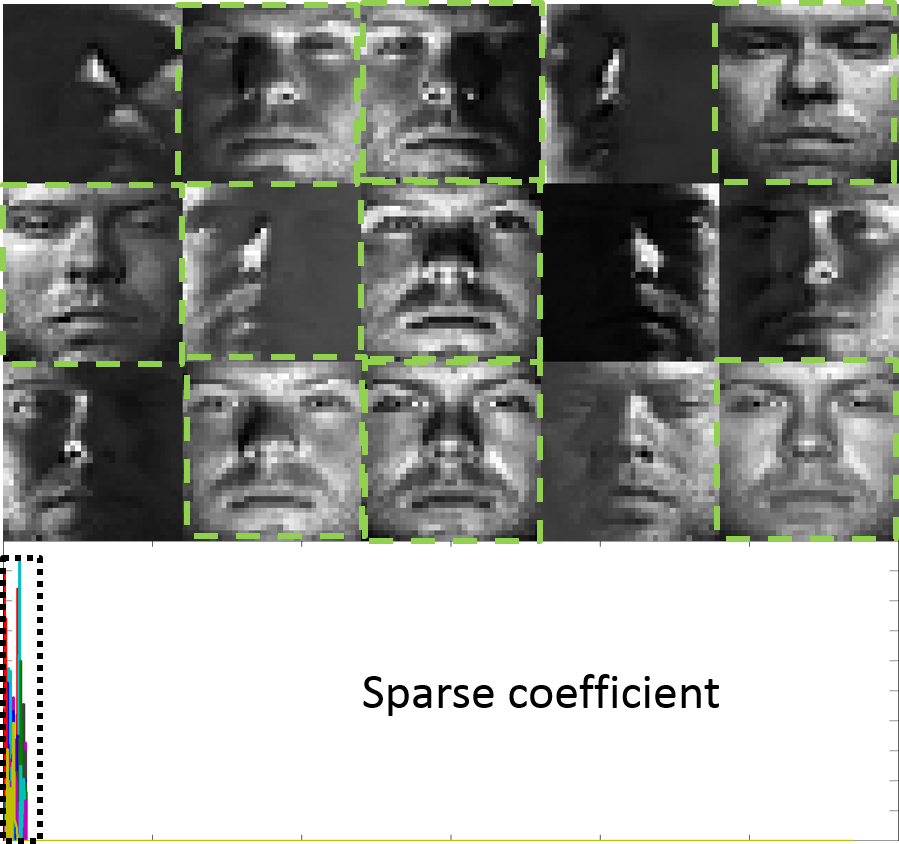}}
\subfigure[K-SVD result (Person 36)]{\includegraphics[width=1.5in]{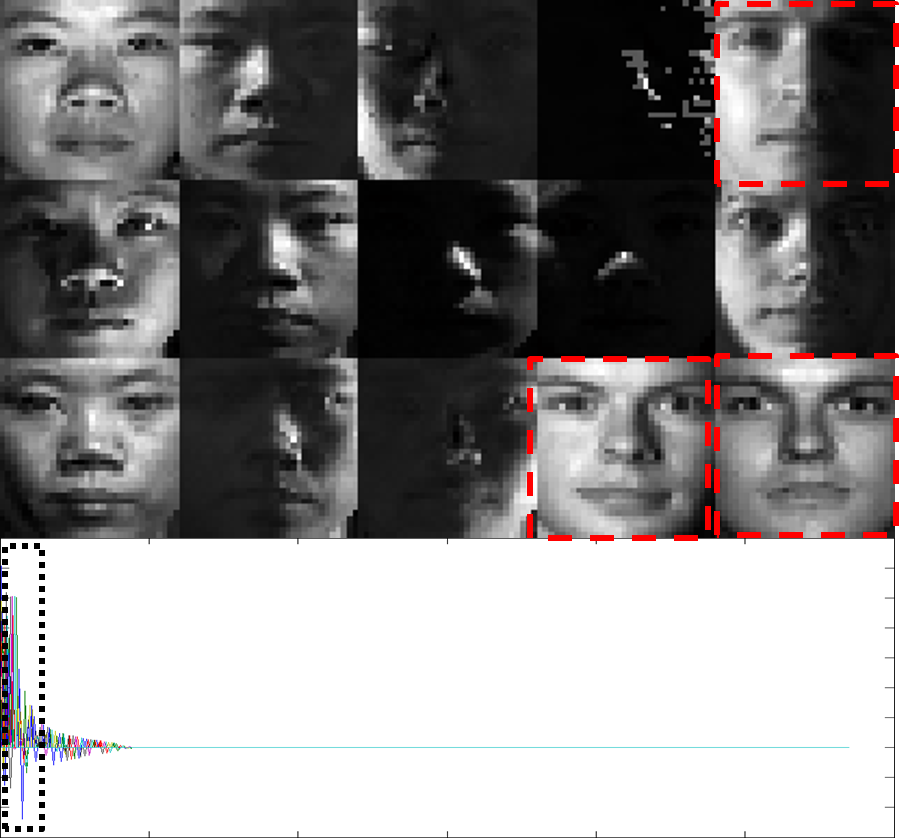}}
\subfigure[GDDL result (Person 36)]{\includegraphics[width=1.5in]{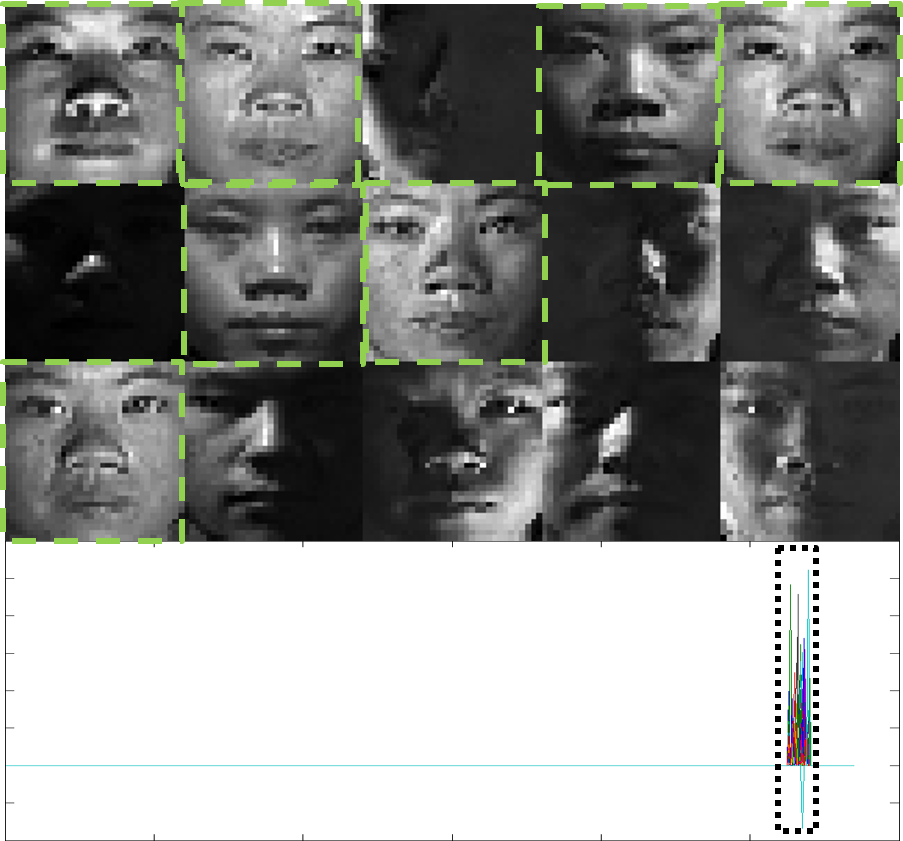}}

\caption{The learned dictionary and the sparse coefficient of training data using K-SVD and GDDL. The sparse codes for all training data in the same class are ploted in the bottom. It can be observed that the labels of dictionary atoms learned by GDDL are consistent while K-SVD can mix the similar faces (red dotted figures). The sparse code for training data indicates that the proposed method can strictly enforce the correct group be chosen while K-SVD fails to do so. Moreover, the dictionary atoms corresponding to the GDDL's shared supports (green dotted figures) capture the similarity between data in the same class while those corresponding to unique supports (un-dotted figures) indicate the within-class variation.}
\label{fig:YaleBDict}
\end{figure*}

The proposed HiDL and GDDL achieve an improvement of more than 3 percentage units in terms of classification accuracy using the same dictionary size for both datasets. To further demonstrate the difference between structured sparsity (i.e., GDDL) and the $l_0$-norm (K-SVD) in DL, the learned dictionary and the sparse code for Person 1 and 36 of Yale B dataset are presented in Fig~\ref{fig:YaleBDict}. Note that the sparse code shown here is that of all training data in each class. The K-SVD dictionary for each class is chosen by finding the dictionary atoms that have the largest magnitude of sparse coefficients, which is same as in Section IV.B. We can see that the K-SVD dictionary has mixed some similar faces from other classes into the desired class (red dotted). Simultaneously, the corresponding sparse code for training data in the same class has a longer-tail distribution outside the group index (Fig~\ref{fig:YaleBDict} (a) and (c)). In contrast, the dictionary learned by GDDL guarantees the dictionary atoms in the group index having the same label. And the sparse code of all training data in this class is strictly within the group index, which justifies our motivation as explained in Fig~\ref{fig:StructDL}. Moreover, the dictionary atoms corresponding to the GDDL's shared supports (green dotted figures in GDDL dictionaries) capture the similarity between data in the same class while those corresponding to unique supports (un-dotted figures in GDDL dictionaries) indicate the within-class variation. 

\begin{table}
\renewcommand{\arraystretch}{1}
\caption{Comparison of proposed HiDL and GDDL with other state-of-art DL methods on face recognition tasks. All methods use the same dictionary size. The best results are achieved by proposed HiDL and GDDL.}
\begin{center}
\begin{tabular}{|c|c|c|c|c|c|c|c|}
\hline
Method & D-KSVD & LLC & LC-KSVD & HiDL & GDDL\\
\hline
\hline
Extended Yale B & 94.1 &90.7  & 95.0 &\textbf{98.0}  &\textbf{98.2}\\
\hline
 AR & 88.8  & 88.7 & 93.7 & \textbf{96.4} &\textbf{96.7}\\
\hline
\end{tabular}
\end{center}
\label{tab:YALEBandAR}
\end{table}

\section{Conclusion}
We incorporate structured sparsity in the DL process for classification purposes. The proposed StructDL framework (including its single task version $-$ HiDL and multi-task version $-$ GDDL) has two advantages compared to $l_0$-and $l_1$-norm regularized methods: \textit{(i)} the dictionary atoms with same group index have same consistent label and this label consistency also exists between dictionary and training data; and \textit{(ii)} the classification performance is more robust to small dictionary size or limited training data, providing computation benefits. Through synthetic and real datasets, we demonstrate that the HiDL and GDDL can generate more discriminative sparse codes, thus improve classification performance. We provide the conditions for HiDL to achieve optimal performance and show the theoretical advantage of HiDL to $l_1$-norm regularized DL for classification tasks. In the future, we will focus on the theoretical analysis of the convergence and locality properties of the proposed HiDL and GDDL. Another interesting direction is to explore the case when the structure is unknown and how to incorporate the learning of structure within the DL process automatically and systematically. 

\bibliographystyle{IEEEtran}
\bibliography{IEEEabrv,IEEEexample}
\end{document}